\documentclass[lettersize,journal]{IEEEtran}
\usepackage{amsmath,amsfonts}
\usepackage{algorithmic}
\usepackage{array}
\usepackage[caption=false,font=normalsize,labelfont=sf,textfont=sf]{subfig}
\usepackage{textcomp}
\usepackage{stfloats}
\usepackage{url}
\usepackage{verbatim}
\usepackage{graphicx}

\usepackage{amsmath}
\usepackage{amssymb}
\usepackage{booktabs}
\usepackage{multirow}
\usepackage{float}
\usepackage{tabularx} 
\usepackage{ragged2e} 
\usepackage{booktabs} 
\usepackage{pifont}
\usepackage{threeparttable}
\usepackage[pagebackref,breaklinks,colorlinks]{hyperref}

\usepackage{colortbl}
\usepackage{color}
\usepackage{xcolor}

\renewcommand{\textcolor}[2]{%
  \ifnum\pdfstrcmp{#1}{blue}=0 
    #2%
  \else%
    \oldtextcolor{#1}{#2}%
  \fi%
}

\def\BibTeX{{\rm B\kern-.05em{\sc i\kern-.025em b}\kern-.08em
    T\kern-.1667em\lower.7ex\hbox{E}\kern-.125emX}}
\usepackage{balance}
\begin{document}
\title{OpenGait: A Comprehensive Benchmark Study for Gait Recognition Towards Better Practicality}
\author{Chao~Fan, 
        Saihui~Hou, 
        Junhao~Liang, 
        Chuanfu~Shen, 
        Jingzhe~Ma,  
        Dongyang~Jin, \\ 
        Yongzhen~Huang, 
        and~Shiqi~Yu
\IEEEcompsocitemizethanks{\IEEEcompsocthanksitem 
Chao Fan is with the School of Artificial Intelligence, Shenzhen University (SZU), the National Engineering Laboratory for Big Data System Computing Technology, SZU, and the Southern University of Science and Technology (SUSTech). E-mail: chaofan996@szu.edu.cn. 

Saihui Hou and Yongzhen Huang are with the School of Artificial Intelligence, Beijing Normal University, Beijing 100875, China, and also with Watrix Technology Limited Co. Ltd, Beijing 100088, China. E-mail: \{housaihui, huangyongzhen\}@bnu.edu.cn.

Junhao Liang is working at the Department of Computer Vision Technology (VIS), Baidu Inc. E-mail: liangjunhao@baidu.com. 

Chuanfu Shen is with the Shenzhen Institute of Advanced Study, University of Electronic Science and Technology of China (UESTC). E-mail: chuanfu.shen@uestc.edu.cn. 

Jingzhe Ma is with the Institute of Applied Artificial Intelligence of the Guangdong-Hong Kong-Macao Greater Bay Area, Shenzhen Polytechnic University. E-mail: jingzhema@szpu.edu.cn. 

Dongyang Jin and Shiqi Yu are with the Department of Computer Science and Engineering, Southern University of Science and Technology (SUSTech), Shenzhen, China. E-mail: 12332451@mail.sustech.edu.cn and yusq@sustech.edu.cn. 

Corresponding author: Shiqi Yu.
\protect\\
}
\thanks{Manuscript received Jan 28, 2024.}
}

\markboth{IEEE Transactions on Pattern Analysis and Machine Intelligence,~Vol.~XX, No.~YY, Month~2024}%
{Fan \MakeLowercase{\textit{et al.}}: OpenGait: A Gait Recognition Benchmark Study Toward Better Practicality}

\maketitle

\begin{abstract}
Gait recognition, a rapidly advancing vision technology for person identification from a distance, has made significant strides in indoor settings. However, evidence suggests that existing methods often yield unsatisfactory results when applied to newly released real-world gait datasets. Furthermore, conclusions drawn from indoor gait datasets may not easily generalize to outdoor ones. Therefore, the primary goal of this paper is to present a comprehensive benchmark study aimed at improving practicality rather than solely focusing on enhancing performance. To this end, we developed OpenGait, a flexible and efficient gait recognition platform. Using OpenGait, we conducted in-depth ablation experiments to revisit recent developments in gait recognition. Surprisingly, we detected some imperfect parts of some prior methods and thereby uncovered several critical yet previously neglected insights. These findings led us to develop three structurally simple yet empirically powerful and practically robust baseline models: DeepGaitV2, SkeletonGait, and SkeletonGait++,  which represent the appearance-based, model-based, and multi-modal methodologies for gait pattern description, respectively. In addition to achieving state-of-the-art performance, our careful exploration provides new perspectives on the modeling experience of deep gait models and the representational capacity of typical gait modalities. 
\textcolor{blue}{
In the end, we discuss the key trends and challenges in current gait recognition, aiming to inspire further advancements towards better practicality.
}
The code is available at \url{https://github.com/ShiqiYu/OpenGait}.
\end{abstract}


\section{Introduction}
\label{sec:intro}
\begin{figure*}[tb]
\centering
\includegraphics[width=0.99\linewidth]{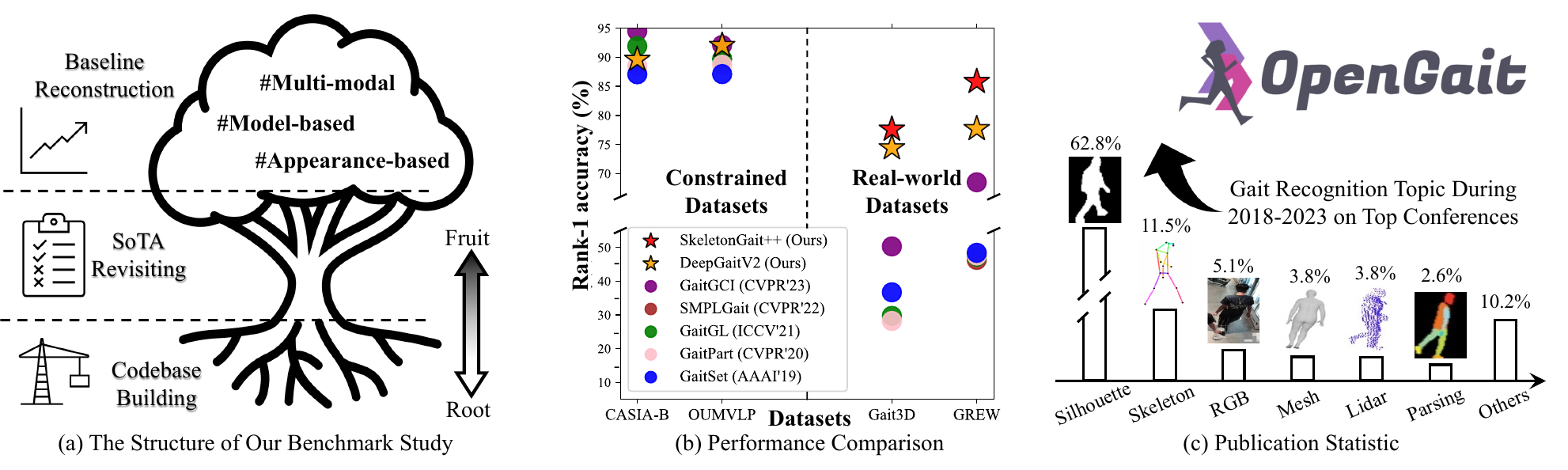}
\caption{
This benchmark study involves the codebase building, previous SoTA revisiting, and strong baseline reconstruction for gait recognition. 
}
\label{fig:intro}
\end{figure*}
\IEEEPARstart{G}{ait} recognition has garnered increasing interest within the vision research community.
It leverages physiological and behavioral characteristics observed in walking videos to authenticate individuals' identities~\cite{wang2003silhouette}.
Compared with other biometric features, such as the face, fingerprint, and iris, 
gait patterns can be captured from a distance in uncontrolled settings, without necessitating physical contact.
Additionally, as a walking behavior, 
gait is inherently difficult to disguise, rendering it theoretically robust against common subject-related variables, such as clothing, carrying items, and posture variations.
These advantages position gait recognition as a promising solution for security applications, including identity authentication and suspect tracking~\cite{wu2016comprehensive}.

The widely acknowledged effectiveness and robustness of deep learning~\cite{ren2015faster, long2015fully, wang2016temporal, cao2017realtime} have greatly propelled various vision techniques forward.
Gait recognition employing deep models has likewise achieved remarkable success~\cite{wu2016comprehensive, chao2019gaitset}.
However, emerging evidence~\cite{zhu2021gait, zheng2022gait3d} suggests that many gait recognition methods cannot perform well enough in practical scenarios.
For instance, as depicted in Fig.~\ref{fig:intro} (b), several representative gait models exhibit a significant accuracy degradation of over 40$\%$ when transitioning from the laboratory testing to the in-the-wild evaluation. 
This performance gap is likely caused by some real-world noisy factors, such as complex occlusions, background variations, and illumination changes.
Through our extensive ablation study, we further identify additional sensitive aspects potentially making some conclusions drawn by prior methods vary across gait datasets. 
Therefore, beyond proposing just another method, the primary objective of this work is to present a comprehensive benchmark study to enhance the practicality of gait recognition. 
To this end, three-fold efforts have been made as illustrated by Fig.~\ref{fig:intro} (a): establishing a solid codebase, exposing gaps by revisiting SoTA methods, and inspiring future research by reconstructing gait baseline models.

In previous studies, the source code was typically shared in the authors' repositories, and the methods relied heavily on in-the-lab gait datasets, particularly CASIA-B~\cite{yu2006framework} and OU-MVLP~\cite{takemura2018multi}. 
However, these datasets have significant limitations compared to in-the-wild datasets. 
For instance, CASIA-B includes data from only 124 subjects collected nearly 20 years ago, while OU-MVLP involves only cross-view changes. 
To accelerate real-world applications, there is a pressing need to build a unified evaluation platform that encompasses various gait methods and emerging gait datasets captured in realistic environments. 
To this end, we introduce OpenGait, a flexible and efficient gait recognition platform, with its highlight features detailed in Sec.~\ref{sec:opengait.1}. 
Thanks to its extensibility and reusability, OpenGait has been extended to some new influential repositories, such as Gait3D-Benchmark~\cite{zheng2022gait3d}\footnote{\url{https://github.com/Gait3D/Gait3D-Benchmark}}, FastPoseGait~\cite{meng2023fastposegait}\footnote{\url{https://github.com/BNU-IVC/FastPoseGait}}, and All-in-One-Gait\footnote{\url{https://github.com/jdyjjj/All-in-One-Gait}}. 
Moreover, OpenGait has also been widely adopted in two major international gait recognition competitions, \textit{i.e.}, HID~\cite{yu2022hid}, and GREW~\cite{zhu2021gait}.
Notably, all top-10 winning teams at HID2022~\cite{yu2022hid} and 2023~\cite{yu2023hid} chose OpenGait as the codebase to develop their solutions.

With the help from OpenGait, 
we efficiently reproduce several SoTA methods, with some results presented in Fig.~\ref{fig:intro} (b).
Furthermore, we reassess some commonly accepted conclusions by re-implementing the ablation study on recently curated outdoor gait datasets. 
Surprisingly, we find that the MGP branch proposed by GaitSet~\cite{chao2019gaitset}, the FConv proposed by GaitPart~\cite{fan2020gaitpart}, the local feature extraction branch proposed by GaitGL~\cite{lin2021gait}, and the SMPL branch proposed by SMPLGait~\cite{zheng2022gait3d}, do not demonstrate superiority on real-world gait datasets. 
With exhaustive analysis, we reveal several critical yet overlooked limitations of existing gait research, including insufficient ablation study, lack of outdoor evaluation, and the absence of a strong backbone, \textit{etc}.

Inspired by the above discoveries, we develop three structurally simple, experimentally powerful, and empirically robust baseline models for real-world gait recognition. 
Specifically, we introduce DeepGaitV2\footnote{We regard GaitBase, proposed by the conference version of this work~\cite{Fan_2023_CVPR}, as an initial exploration of deep ResNet~\cite{he2016deep} for gait modeling and denote it as DeepGaitV1. Consequently, its subsequent iteration is labeled as DeepGaitV2.}, SkeletonGait, and SkeletonGait++, each utilizing the silhouette image, skeletal coordinates, and the combination of these two as input, accordingly standing for the appearance-based, model-based, and multi-modal methodology for gait pattern description.
As illustrated in Fig.~\ref{fig:intro} (c), the silhouette and skeleton present two of the most popular gait inputs\footnote{We count 78 gait papers released over the past 5 years, across top computer vision conferences such as CVPR, ICCV, ECCV, BMVC, WACV, ACCV, and related top conferences like AAAI, ACM MM, ICASSP,  ICIP, and IJCB.} underscoring their adoption. 
In terms of network design, this benchmark study prioritizes the use of widely accepted modules and gait frameworks to enhance the practicality and applicability of our findings. 
Through extensive experiments, several critical insights regarding the modeling capacity of shallow \textit{vs.} deep gait models and the representational ability of gait silhouette \textit{vs.} skeleton has been provided. 
By addressing above challenges straightly, our pragmatic solutions yield significant advancements as depicted in Fig.~\ref{fig:intro} (b). 
More details will be provided in Sec.~\ref{sec:gait_model}. 

\textcolor{blue}{
Before the conclusion, this study presents a concise discussion on key trends and challenges in gait recognition under real-world scenarios.
This section covers critical topics such as ideal gait input, handling noisy data, in-the-wild challenges, gait metric learning, and gait temporal modeling. By offering valuable insights into the current state of gait recognition and outlining a clear roadmap for future advancements, it aims to drive research toward better practicality.
}

In summary, this benchmark study makes four key contributions: 
a) We develop OpenGait, a unified and extensible platform designed to facilitate the systematic analysis of gait recognition.
b) We conduct a comprehensive experimental review of representative gait recognition methods, yielding insightful findings. 
c) We introduce three robust baseline models, while maintaining practicality and commonality, demonstrating significant performance enhancements.
\textcolor{blue}{
d) We provide a brief discussion section highlighting key trends and challenges to inspire future gait recognition research.
}

This work stems from our CVPR 2023 highlight paper, OpenGait~\cite{Fan_2023_CVPR}, and makes four significant advancements: 
a) We upgrade the GaitBase to DeepGaitV2 by thoroughly exploring the capability of shallow \textit{vs.} deep models for gait modeling.
b) We combine findings from our AAAI 2024 publication~\cite{fan2024skeletongait}, specifically SkeletonGait and SkeletonGait++, with DeepGaitV2, resulting in a more comprehensive baseline reconstruction study. 
c) We broaden the experimental scope to include recently released popular gait datasets, such as CCPG~\cite{Li_2023_CVPR} and SUSTech1K~\cite{Shen_2023_CVPR}. 
\textcolor{blue}{
(d) We add a brief section discussing the key trends and challenges in gait recognition, incorporating feedback from the OpenGait community.
}
Based on these significant changes, this paper offers a more systematic analysis aimed at improving the practicality of gait recognition compared to its conference counterpart. 

\section{Related Work}
This section starts with a literature review that covers the gait datasets, input modalities, and popular methods. It aims to provide a snapshot of recent developments in the field. 

\begin{table*}[tb]
\centering
\caption{
The Comparison Between Six Popular Gait Datasets. 
}
\renewcommand{\arraystretch}{1.2}
\begin{threeparttable}
\begin{tabular}{c|ccccc|c|ccc}
\toprule
\multirow{2}{*}{Environment} & \multirow{2}{*}{Dataset}   & \multicolumn{2}{c}{Train Set} & \multicolumn{2}{c|}{Test Set} & \multirow{2}{*}{Cameras} & \multirow{2}{*}{Variations}  & \multirow{2}{*}{Modalities}               & \multirow{2}{*}{\textcolor{blue}{Year}} \\ 
& & \#ID            & \#Seq           & \#ID           & \#Seq           &     & & \\ \hline \hline

\multirow{7}{*}{Constrained} & CASIA-B~\cite{yu2006framework}   & 74            & 8,140         & 50           & 5,500    & 11      & \#CV, \#CL, \#BG                                                                          & Sil., RGB                      & \textcolor{blue}{2006}     \\
                             & OU-MVLP~\cite{takemura2018multi}   & 5,153         & 144,284       & 5,154        & 144,412   & 14      & \#CV                                                                                              & Sil., Ske.                  & \textcolor{blue}{2018}     \\
                             & \textcolor{blue}{FVG~\cite{zhang2019gait}} & \textcolor{blue}{136}         & \textcolor{blue}{-}       & \textcolor{blue}{90}        & \textcolor{blue}{-}    & \textcolor{blue}{1}      & \textcolor{blue}{\#CV, \#CL, \textit{etc}} & \textcolor{blue}{RGB}       & \textcolor{blue}{2019}     \\ 
                             & CCPG~\cite{Li_2023_CVPR}      & 100            & 8,187         & 100           & 8,095    & 10      & \#CV, \#CL, \#BG                                                                          & Sil., RGB                      & \textcolor{blue}{2023}     \\
                             & SUSTech1K~\cite{Shen_2023_CVPR} & 200         & 5,988       & 850        & 19,228    & 12      & \#CV, \#CL, \#BG, \textit{etc} & Sil., RGB, Lidar       & \textcolor{blue}{2023}     \\ 
                             & \textcolor{blue}{CCGR~\cite{zou2024cross}} & \textcolor{blue}{600}         & \textcolor{blue}{908322}       & \textcolor{blue}{370}        & \textcolor{blue}{672295}    & \textcolor{blue}{33}      & \textcolor{blue}{\#CV, \#CL, \#BG, \textit{etc}} & \textcolor{blue}{Sil., RGB, Parsing, \textit{etc}}     & \textcolor{blue}{2024}     \\ 
                             \midrule
\multirow{2}{*}{In-the-wild} & GREW~\cite{zhu2021gait}      & 20,000        & 102,887       & 6,000        & 24,000   & 882     & Real-world                                                                                         & Sil., Ske., Flow & \textcolor{blue}{2021}     \\
                             & Gait3D~\cite{zheng2022gait3d}    & 3,000         & 18,940        & 1,000        & 6,369    & 39      & Real-world                                                                                         & Sil., Ske., Mesh   & \textcolor{blue}{2022}     \\ \bottomrule
\end{tabular}
\textit{Note: \#ID and \#Seq present the number of identities and sequences. \#CV, \#CL, and \#BG are for camera viewpoints, clothing changes, and bag carrying. Sil. and Ske. respectively refer to the silhouette and skeleton.} \\
\end{threeparttable}
\label{tab:gait_datasets} 
\end{table*}

\subsection{Gait Recognition Datasets}
The large-scale data collection presents an essential premise for gait recognition research. 
As exhibited in Table~\ref{tab:gait_datasets}, several popular gait datasets can be roughly categorized into two groups: the constrained and the in-the-wild ones. 
The former group often requires subjects to repeatedly walk along fixed routes with simulating real scenarios by introducing dressing and carrying changes. 
The latter ones should be captured from real-world scenarios and cover a wide range of real-world covariates. 
Even though the constrained gait datasets can include many human-crafted complexities, 
in-the-wild data is still essential to gait recognition~\cite{shen2022comprehensive}. 
Before the release of this paper's conference version~\cite{Fan_2023_CVPR}, many existing works only verify their effectiveness on constrained in-the-lab datasets, \textit{e.g.}, CASIA-B~\cite{yu2006framework} and OU-MVLP~\cite{takemura2018multi}, thus posing a high risk of vulnerability for practical usage. 
In this benchmark study, we evaluate several representative methods on emerging real-world gait datasets thus uncovering overlooked issues. 
In addition, the effectiveness of our three baseline models has been verified on both constrained datasets and in-the-wild ones. 
The results have convincingly demonstrated the generality and significance of our findings for gait recognition. 

\textcolor{blue}{
Beyond identification purposes, gait datasets for medical applications have gained increasing attention in recent literature. 
Notable examples include Scoliosis1K~\cite{zhou2024gait} for scoliosis classification, as well as PsyMo~\cite{cosma2024psymo} and D-Gait~\cite{liu2024depression} for psychological assessment. 
OpenGait has integrated support for some of these datasets and recognizes vision-based gait analysis as a promising direction for medical research.
}

\subsection{Typical Gait Modalities}
As shown in Fig.~\ref{fig:intro} (c), 
the mainstream gait modalities, such as the binary silhouette, skeleton coordinates, RGB image, human mesh, and body parsing image, are primarily derived from cameras and represented by various data formats.
During this process, diverse pretreatment operations~\cite{chao2019gaitset, liao2020model, song2019gaitnet, Shen_2023_CVPR, zheng2023parsing} and end-to-end learning manners~\cite{liang2022gaitedge, li2020end} have been utilized to mitigate the influence of gait-irrelevant noises like color, texture, and background cues. 
Out of this trend, some studies propose novel gait modalities by incorporating emerging sensors such as LiDAR~\cite{Shen_2023_CVPR} and event cameras~\cite{event2022}. 
However, these sensors have not been widely deployed in surveillance scenarios, making them unsuitable for large-scale surveillance applications currently. 
To ensure comprehensiveness, OpenGait supports all of the aforementioned gait modalities, regardless of sensor types.

In addition, Fig.~\ref{fig:intro} (c) further reveals that binary silhouettes and skeleton coordinates act as two of the most prevailing gait modalities in the latest literature. 
They both explicitly present the human body structural characteristics, \textit{e.g.}, the length, ratio, and movement of human limbs.
For the rich body shape information, the silhouette has a better discriminative capacity in gait description. 
However, regardless of the modality employed, 
gait recognition methods will suffer accuracy degradation when tested in outdoor scenarios.~\cite{Fan_2023_CVPR, Fu_2023_ICCV}. 
Beyond performance, we also try to explore the cooperativeness and complementarity natures of these two gait modalities by proposing a multi-modality method, SkeletonGait++.

\subsection{Gait Recognition Methods}
For convenience, we roughly group gait recognition methods into three categories, \textit{i.e.}, the model-based, appearance-based, and multi-modal methods. 

\noindent
\textbf{Model-based Gait Recognition} methods~\cite{liao2020model,teepe2021gaitgraph,li2020end, li2022cyclegait} tend to take the estimated underlying structure of the body as input, such as 2D/3D pose and SMPL~\cite{loper2015smpl} model.
With extremely excluding visual clues, these gait modalities, typically parameterized as coordinates of body joints or customized vectors, are ideally \textit{clean} against factors like carrying and dressing items. 
In the recent literature, PoseGait~\cite{liao2020model} employs 3D body pose and hand-crafted structural features to overcome the changes in clothing.
GaitGraph~\cite{teepe2021gaitgraph} introduces the graph convolutional network for 2D skeleton-based gait description. 
In GPGait~\cite{Fu_2023_ICCV} a human-oriented transformation and a series of human-oriented descriptors are developed to generate the unified pose representation with multiple features. 
\textcolor{blue}{GaitPT~\cite{catruna2024gaitpt} adopts a hierarchical transformer design,  effectively extracting spatial and temporal movement features in an anatomically consistent manner.}
Generally, the skeleton-based methods often struggle with the absence of body shape and, as a result, usually perform unsatisfactorily on most gait datasets.
On the other hand, several SMPL-based methods have achieved significant advances on indoor OU-MVLP~\cite{takemura2018multi}, \textit{e.g.}, Li~\textit{et al.}~\cite{li2020end} fine-tuned a pre-trained human mesh recovery network to construct the end-to-end SMPL-based model and Xu~\textit{et al}.~\cite{xu2023occlussion} proposed an occlusion-aware human mesh based method to alleviate the partial occlusion problem within practical gait recognition. 
However, model-based methods on real-world gait datasets~\cite{zhu2021gait, zheng2022gait3d} have not yet been verified exactly. 

\noindent
\textbf{Appearance-based Gait Recognition} methods mostly learn gait features from binary silhouettes or RGB images and can benefit from informative shape characteristics. 
With the boom of deep learning, 
most current appearance-based works focus on spatial feature extraction and temporal modeling.
Specifically, 
in GaitSet~\cite{chao2019gaitset} a gait sequence is innovatively regarded as a set and the max pooling is utilized to compress frame-level spatial features. 
Thanks to its simplicity and effectiveness, 
GaitSet has become one of the most influential gait methods in recent years.
In GaitPart~\cite{fan2020gaitpart} the local details of the input silhouette are carefully explored and  the temporal dependencies are modeled.
In GaitGL~\cite{lin2021gait} it is argued that the spatially global gait representations often neglect the details, and the local region-based descriptors cannot capture the relations among neighboring parts.
3DLocal~\cite{huang20213d} tries to extract limb features through 3D local operations at adaptive scales, and 
DyGait~\cite{wang2023dygait} tries to establish spatial-temporal representations of dynamic body parts. 
Generally speaking, appearance-based methods contribute the majority in the recent gait recognition.

\noindent
\textbf{Multi-modal Gait Recognition} methods typically incorporate multiple gait modalities as input, reflecting an emerging trend in gait pattern description. 
In recent literature, SMPLGait~\cite{zheng2022gait3d} exploits 3D information from the SMPL model to enhance the learning of gait appearance features. 
BiFusion~\cite{peng2021learning} integrates body skeletons and silhouettes to extract the rich spatio-temporal features.
ParsingGait~\cite{zheng2023parsing} regards a human parsing image as a form of fine-grained segmentation to improve the representational ability of its silhouette branch. 

\begin{figure*}[tb]
\centering
\includegraphics[height=3.5cm]{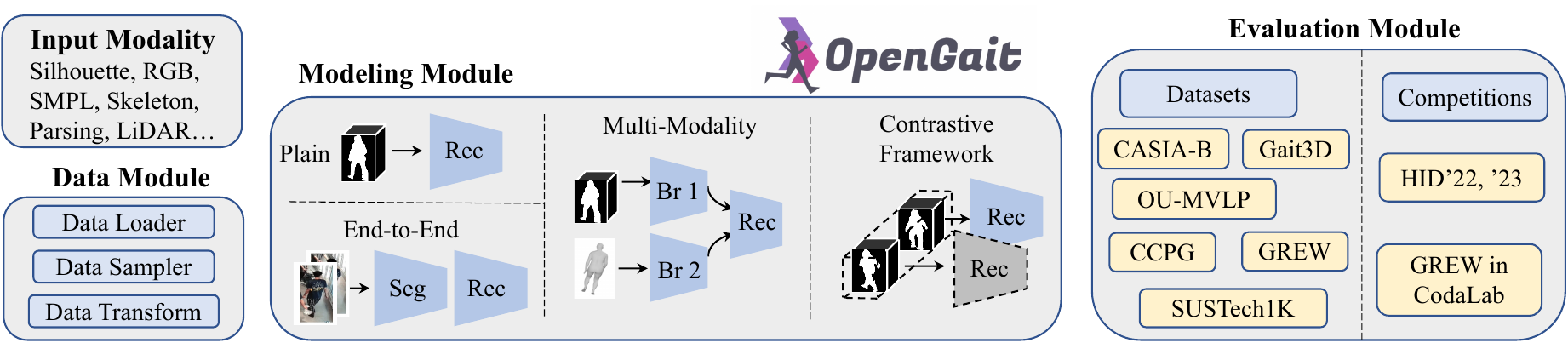}
\caption{The three main modules of \textbf{OpenGait}. In the Modeling Module, 
Seg is for \textit{segmentation}, 
Rec is for \textit{recognition}, 
and Br is for \textit{branch}.
}
\label{fig:framework}
\end{figure*}

In addition to the above, 
many successful model-agnostic gait frameworks promote gait research as well. 
For example, 
GaitEdge~\cite{liang2022gaitedge} employs edge-trainable silhouettes to build an end-to-end gait recognition framework, 
GaitGCI~\cite{dou2023gaitgci} introduces a generative counterfactual intervention to alleviate the over-fitting problem within gait recognition, 
GaitSSB~\cite{fan2022learning} builds a pre-trained model with millions of unlabelled gait sequences by contrastive learning, 
LidarGait~\cite{Shen_2023_CVPR} collects the first large-scale in-the-wild lidar-based gait dataset to explore gait feature extraction from precise 3D point clouds, 
and UDA~\cite{ma2023fine} develops a novel fine-grained unsupervised domain adaptation method for robust gait recognition. 

To be noting that almost all of the aforementioned methods and some other recent methods have been implemented in OpenGait. OpenGait can help researchers to evaluate and compare different methods efficiently.

\subsection{Other Related Works}
\noindent
\textbf{Survey on Gait Recognition.}
While survey papers~\cite{filipi2022gait,sepas2022deep,shen2022comprehensive} have offered comprehensive overviews on gait recognition, detailed experimental analysis and comparisons are still lacking. 
In contrast, several impactful works in other fields, such as recommendation system~\cite{ferrari2019we}, metric learning~\cite{musgrave2020metric}, and unsupervised domain adaptation~\cite{musgrave2021unsupervised}, have revisited SoTA experimentally. 
Similarly, OpenGait aims to do this by not only re-evaluating recent methods but also uncovering some new insights in gait recognition.

\noindent
\textbf{Codebase for Computer Vision.}
In the computer vision community, a robust codebase is crucial for advancing research in specific domains.
For instance, Amos \textit{et al.}~\cite{amos2016openface} proposes OpenFace, a face recognition library that bridges the gap between public face recognition systems and industry-leading private systems. 
In object detection, mmdetection~\cite{mmdetection} supports almost all popular detection methods, 
providing a convenient platform for systematic comparison.
As gait recognition continues to evolve rapidly, 
the need for an infrastructure code platform has become increasingly evident. 

\section{A Glance at OpenGait Platform}
\label{sec:opengait}
Over the past few years, numerous new methods and datasets have emerged for gait recognition. 
However, the lack of a unified and fair evaluation platform cannot be overlooked.
A PyTorch-based~\cite{paszke2019pytorch} toolbox, termed OpenGait, is presented in the following as a solution to this issue.

\subsection{Design Principles of OpenGait}
\label{sec:opengait.1}

We have considered the following design principles below when we designed OpenGait.

\noindent
\textbf{Compatibility with Multiple Gait Modalities.} 
Typical gait modalities include silhouette images, 2D/3D skeletons, and emerging ones like the SMPL model~\cite{li2020end,zheng2022gait3d}, point clouds~\cite{Shen_2023_CVPR}, and RGB images~\cite{song2019gaitnet,liang2022gaitedge}. 
Most open-source projects are designed for specific methods and mostly supporting just one or a limited number of  gait modalities.
In this work, OpenGait tries to support all popular modalities.

\noindent
\textbf{Compatibility with Various Frameworks.} 
The landscape of gait recognition is evolving with the emergence of novel frameworks, such as multi-modal~\cite{zheng2022gait3d}, end-to-end~\cite{song2019gaitnet,li2020end,liang2022gaitedge}, and contrastive learning~\cite{fan2022learning} based methods. 
Unlike some open-source projects specifically tailored to certain models, OpenGait is designed to be flexible and extensible, and can accommodate all of the above frameworks seamlessly.

\noindent
\textbf{Support for Various Evaluation Datasets.} 
OpenGait offers compatibility with commonly used gait datasets, which cover a wide range of scenarios and sensors. 
It seamlessly integrates with indoor datasets like CASIA-B~\cite{yu2006framework} and OU-MVLP~\cite{takemura2018multi}, as well as in-the-wild datasets, such as GREW~\cite{zhu2021gait} and Gait3D~\cite{zheng2022gait3d}. 
Moreover, OpenGait also supports the latest in-the-lab gait datasets such as CCPG~\cite{Li_2023_CVPR} and SUSTech1K~\cite{Shen_2023_CVPR}.
Thanks to its flexibility and reusability, OpenGait has been extended to major gait recognition competitions like HID~\cite{yu2022hid} and GREW~\cite{zhu2021gait}, where it has been instrumental in the development of winners' solutions.

\noindent
\textbf{Support for SoTA.} 
OpenGait encompasses the reproduction of most SoTA methods mentioned and will keep updated to date by appending promising methods.
The results by the reproduced methods match or surpass the ones reported in the original papers. 
In this way, we provided rich examples to help beginners get started with the code. 
Moreover, OpenGait also presents a solid platform for fair and thorough comparisons in this benchmark study.

\subsection{Modules in OpenGait}
\label{sec:modules}
We follow the designs of a typical PyTorch deep learning project and divide OpenGait into three modules, \textit{i.e.}, \textit{data}, \textit{modeling}, and \textit{evaluation}, as shown in Fig.~\ref{fig:framework}. 

\label{sec:doub}

\textbf{Data module}
contains some data loaders, data samplers, and data transform scripts, and they are for loading, sampling, and pre-processing the input data, respectively.

\textbf{Modeling module}
is built on top of a base class (\texttt{BaseModel}) that pre-defines typical behaviors of the deep model during the training and testing phases, including optimization and inference.
The four essential components of current gait recognition methods, namely, \textit{backbone}, \textit{neck}, \textit{head}, and \textit{loss}, can be customized in this class.

\textbf{Evaluation module}
is used to assess the performance of the obtained model. 
Given that different datasets often come with various evaluation protocols, we integrate them into OpenGait to relieve researchers from dealing with these tedious details. 

OpenGait was developed in Python language, and easy to use. Users can easily  train a new deep model on multiple datasets by modifying some configurations. Its GitHub repository\footnote{\url{https://github.com/ShiqiYu/OpenGait}} has gained 659 stars and 153 forks up to June 2024. 

\section{Revisiting of State-of-the-art Methods}
\label{sec:revisit}
With the help of OpenGait, we can conduct a comprehensive re-evaluation of several representative gait methods efficiently.
Through meticulous ablation studies, some interesting insights, which are different from the ones presented in the original papers, have been recovered. 

\subsection{Experimental Recheck on Representative Methods}
A lot of works only perform experiments on indoor gait datasets, notably CASIA-B~\cite{yu2006framework} and OU-MVLP~\cite{takemura2018multi}, with further ablation studies typically limited on CASIA-B~\cite{yu2006framework}.
This subsection aims to extend some additional ablation studies to a large in-the-wild dataset, Gait3D~\cite{zheng2022gait3d}, to assess the methods' robustness in real-world scenarios.

\begin{table}[htbp]
\centering
\caption{The effectiveness of MGP and Multi-scale HPP in GaitSet~\cite{chao2019gaitset}.}
\renewcommand{\arraystretch}{1.2}
\resizebox{0.90\columnwidth}{!}{
\begin{tabular}{c|c|ccc|cc} 
\toprule
\multirow{2}{*}{MGP}      & \multirow{2}{*}{Multi-scale HPP} & \multicolumn{3}{c|}{CASIA-B}                  & \multicolumn{2}{c}{Gait3D}                       \\ 
\cline{3-7}
                          &                              & NM            & BG            & CL            & R-1                    & R-5                     \\ 
\hline\hline
\multirow{2}{*}{\ding{52}}         & \ding{56}                           & \textbf{95.9} & 90.3          & \textbf{74.2} & 44.3                   & 64.7                    \\ 
\cline{2-7}
                          & \ding{52}                           & 95.8          & 90.4          & 73.2          & 44.3                   & 64.4                    \\ 
\hline
\multirow{2}{*}{\ding{56}} & \ding{56}                           & 95.3          & \textbf{90.5} & 74.0          & \textbf{\textbf{45.8}} & \textbf{\textbf{65.1}}  \\ 
\cline{2-7}
                          & \ding{52}                           & 94.5          & 89.1          & 72.3          & 43.7                   & 63.8                    \\
\bottomrule
\end{tabular}
}
\label{tab:mgp}
\end{table}

\begin{figure*}[tb]
\centering
\includegraphics[height=6.5cm]{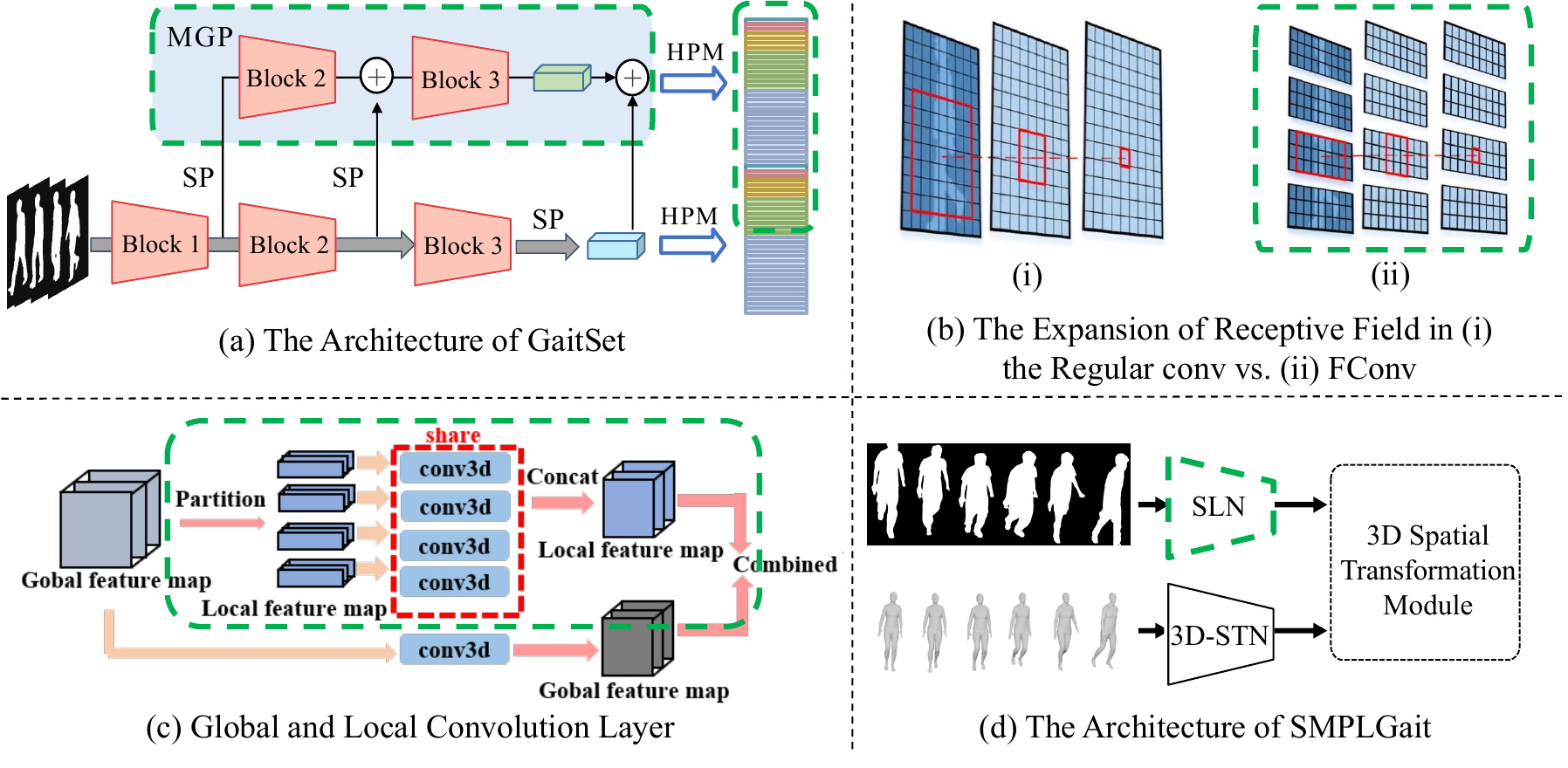}
\caption{The key modules of four SoTA methods. 
The modules enclosed by dotted green lines are the ones  removed or replaced for comparisons. 
(a): MGP and multi-scale mechanism in HPP are to be removed~\cite{chao2019gaitset}.
(b): FConvs are to be replaced with the regular convolution layers~\cite{fan2020gaitpart}. 
(c): The upper local feature branch is to be replaced with an independent conv3d that is identical to the lower global branch~\cite{lin2021gait}. 
(d): SLN branch is to be replaced with our strong backbone~\cite{Fan_2023_CVPR}.}
\label{fig:flaws}
\end{figure*}

\noindent \textbf{Re-conduct  Ablation Study of GaitSet.}
With taking the silhouettes as input, 
GaitSet~\cite{chao2019gaitset} treats the gait sequence as an unordered set and uses a simple maximum pooling function along the temporal dimension, called Set Pooling (SP), to extract a set-level understanding of the entire input video.
GaitSet~\cite{chao2019gaitset} provides insights for many subsequent works for its simplicity and effectiveness. 
However, we find that two important components in GaitSet~\cite{chao2019gaitset}, 
namely the parallel Multi-layer Global Pipeline (MGP) and pyramid-like Horizontal Pyramid Pooling (HPP)~\cite{fu2019horizontal},
do not work well enough on the indoor CASIA-B and outdoor Gait3D datasets. 
Specifically, as shown in Fig.~\ref{fig:flaws} (a), 
MGP can be regarded as an additional branch that aggregates the hierarchical set-level characteristics. HPP~\cite{fu2019horizontal} follows the fashion of feature pyramid structure, and tries to extract multi-scale part-based features. 
As shown in Table~\ref{tab:mgp}, if we strip the MGP or remove the multi-scale mechanism in HPP from the original GaitSet~\cite{chao2019gaitset}, the revised model can reach the same or even better performance on both CASIA-B and Gait3D with saving over 80\% training weights.
This result indicates that the set-level characteristics extracted by the bottom convolution blocks may be difficult to benefit the final gait representation\footnote{
In the original GaitSet~\cite{chao2019gaitset}, all the obtained partial vectors will be concatenated into a single feature vector used for final evaluation. However, the follow-up works~\cite{fan2020gaitpart, lin2021gait, zheng2022gait3d} find this concatenation unnecessary and take the partial distance as a final metric. Therefore, OpenGait follows this manner and reproduces a better performance than the original GaitSet.
}.
Besides, 
the multi-scale mechanism in HPP cannot provide discriminative features. 
The cause may be that the employed statistical pooling functions are too weak to learn extra knowledge from various-scale human body parts.

\begin{table}[t]
\centering
\caption{The effect of FConv in GaitPart~\cite{fan2020gaitpart}. }
\renewcommand{\arraystretch}{1.2}
\begin{tabular}{c|ccc|cc} 
\toprule
\multirow{2}{*}{FConv} & \multicolumn{3}{c|}{CASIA-B}         & \multicolumn{2}{c}{Gait3D}  \\ 
\cline{2-6}
                        & NM            & BG            & CL            & R-1           & R-5                  \\ 
\hline\hline
\ding{52}                & \textbf{96.2} & \textbf{91.5} & \textbf{78.7} & 29.2          & 48.6                 \\ 
\hline
\ding{56}               & 95.6          & 88.4          & 76.1          & \textbf{36.2} & \textbf{57.0}        \\
\bottomrule
\end{tabular}
\label{tab:fconv}
\end{table}

\noindent \textbf{Re-conduct  Ablation Study of GaitPart.} 
One of the core contributions of GaitPart~\cite{fan2020gaitpart} is to point out the importance of local details with the proposed Focal Convolution (FConv) layer. 
Fig.~\ref{fig:flaws} (b) shows the receptive field's expansion of the top-layer neuron in the network established by regular \textit{vs.} focal convolution layers.
Technically, 
FConv splits the input feature map into several parts horizontally and then performs a regular convolution over each part separately. 
As shown in Table~\ref{tab:fconv}, we get much better performance (Rank-1: +7.0\%) on Gait3D by degenerating FConv into the regular convolution layer.
This phenomenon exhibits that the extraction of gait features may be seriously affected by merely splitting the feature map in a fixed-window manner, due to the low-quality segmentation of the outdoor data.
The shifted window mechanism~\cite{liu2021swin} may offer a better solution to this issue.

\begin{table}[htbp]
\centering
\caption{The effect of local branch in GaitGL~\cite{lin2021gait}.}
\renewcommand{\arraystretch}{1.2}
\resizebox{.75\columnwidth}{!}{
\begin{tabular}{c|ccc|cc} 
\toprule
\multirow{2}{*}{Local Branch} & \multicolumn{3}{c|}{CASIA-B}         & \multicolumn{2}{c}{Gait3D}  \\ 
\cline{2-6}
                        & NM            & BG            & CL            & R-1           & R-5                  \\ 
\hline\hline
\ding{52}       & \textbf{97.4} & \textbf{94.5} & \textbf{83.6} & 31.4          & 50.0                 \\ 
\hline
\ding{56}      & 97.1          & 93.7          & 81.9          & \textbf{32.2} & \textbf{52.5}        \\
\bottomrule
\end{tabular}
}
\label{tab:glfe}
\end{table}

\noindent \textbf{Re-conduct  Ablation Study of GaitGL.}
As shown in Fig~\ref{fig:flaws} (c), GaitGL~\cite{lin2021gait} develops the global and local convolution layer, where the local branch can be regarded as the FConv~\cite{fan2020gaitpart} employing 3D convolution while the global branch presents a standard 3D convolution layer.
Similar to GaitPart~\cite{fan2020gaitpart}, 
as shown in Table~\ref{tab:glfe},
removing the local branch can achieve better performance on the Gait3D dataset.

\noindent \textbf{Re-conduct  Ablation Study of SMPLGait.}
As shown in Fig.~\ref{fig:flaws} (d), 
SMPLGait~\cite{zheng2022gait3d} consists of two elaborately-designed branches,
\textit{i.e.}, the silhouette (SLN) and SMPL (3D-STN) branch, respectively using for 2D appearance extraction and 3D knowledge learning. 
SMPLGait~\cite{zheng2022gait3d} takes advantage of the 3D mesh data available in Gait3D~\cite{zheng2022gait3d} and achieves performance gains on the top of the silhouette branch.  
However, Table~\ref{tab:smpl} demonstrates that the proposed SMPL branch does not provide obvious benefits when we give the silhouette branch a strong backbone network like GaitBase~\cite{Fan_2023_CVPR}.

\begin{table}[t]
\centering
\caption{The effect of SMPL branch in SMPLGait~\cite{zheng2022gait3d}. }
\renewcommand{\arraystretch}{1.2}
\begin{threeparttable}
\begin{tabular}{c|cc|cc}
\toprule
\multirow{3}{*}{SMPL branch} & \multicolumn{4}{c}{Silhouette branch}                                     \\ 
\cline{2-5}
                        & \multicolumn{2}{c|}{SLN} & \multicolumn{2}{c}{ResNet9}  \\ 
\cline{2-5}
                        & R-1           & R-5               & R-1           & R-5                   \\ 
\hline\hline
\ding{52}                & \textbf{46.3} & \textbf{64.5}     & 55.2          & \textbf{75.7}         \\ 
\hline
\ding{56}  & 42.9         & 63.9             & \textbf{56.5} & 75.2                  \\
\bottomrule
\end{tabular}
\textit{Note: SLN and ResNet9 denote the backbone used by SMPLGait~\cite{zheng2022gait3d} and GaitBase~\cite{Fan_2023_CVPR}, respectively.}
\end{threeparttable}
\label{tab:smpl}
\end{table}

In our view, there should be three potential reasons causing the failure of the SMPL branch: a) 
Though the SMPL model is usually visualized as a dense mesh, its feature vector only possesses tens of dimensions that present the relatively sparse characterization of body shape and posture, making it challenging to enhance the fine-grained description of gait patterns.
b) Since the SMPL model is not recognition-oriented, purposefully fine-tuning it may be more optimal than directly utilizing it to depict the subtle individual characteristics as in~\cite{li2020end}. c) In the wild, estimating an accurate SMPL model that finely captures body shape and posture from a single RGB camera is still challenging. In a nutshell, introducing 3D geometrical information from the SMPL model to enhance gait representation learning is well worth further exploration.

\subsection{Rethinking and Discussion}
Drawing from the aforementioned findings, we argue that the deep models in previous gait recognition methods may not be robust enough. 
It is caused by the heavy reliance on constrained datasets that struggle with simulating the complexity of outdoor environments. 
To provide a comprehensive understanding, we elaborate on the following points.

\label{sec:weak_dataset}
\noindent \textbf{Necessity of Outdoor Evaluation.}
Previous methods are primarily evaluated on the in-the-lab CASIA-B~\cite{yu2006framework} and OU-MVLP~\cite{takemura2018multi}.
We argue that this practice suffers from three drawbacks: 
a) \textit{Indoor settings.} 
The walking videos were captured by a camera array, and the subjects were requested to follow a particular route. 
It makes the data significantly different from that from real-world scenarios.
b) \textit{Simple background}.
The simple laboratory background cannot reflect the complex changes of wild scenes.
c) \textit{Outdated processing methods}. 
The silhouettes were obtained by background subtraction algorithms significantly distinct from the deep-learning-based segmentation methods used for current applications.

Recently, some large-scale real-world gait datasets, such as  GREW~\cite{zhu2021gait} and Gait3D~\cite{zheng2022gait3d}, aim to advance gait recognition from controlled laboratory settings to real scenarios.
But designing gait recognition methods based on real-world datasets has still not gained enough attention before the publication of this paper's conference version.

\begin{figure*}[t]
\centering
\includegraphics[height=4cm]{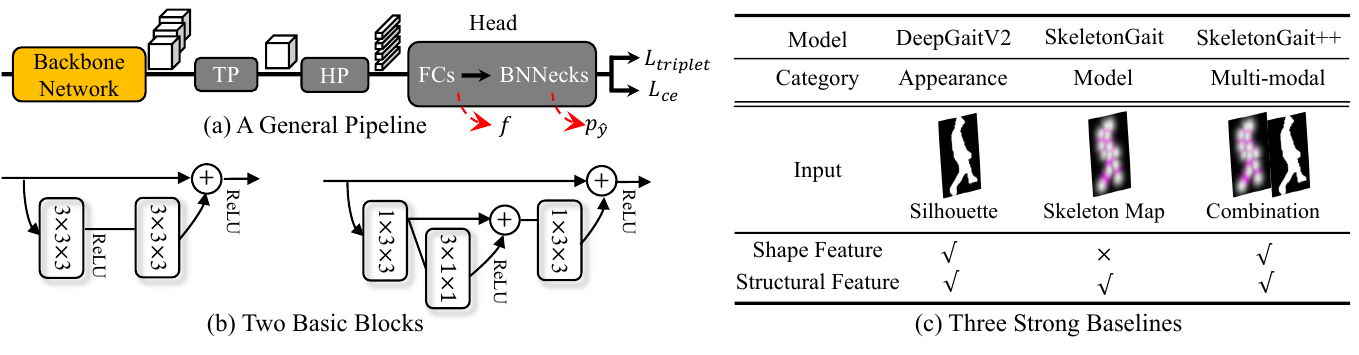}
\caption{
(a) Overall architecture. The TP and HP respectively denote Temporal and Horizontal Pooling. The Head comprises the separate fully-connected and BNNeck~\cite{luo2019bag} layers, respectively outputting the identity feature vector $f$ and identity distribution vector $p_{\hat{y}}$. (b) Employed blocks, \textit{i.e.}, the 3D and Pseudo 3D residual convolution layers~\cite{he2016deep, hara2017learning, qiu2017learning}.
(c) The proposed three baseline models. 
}
\label{fig:pipeline}
\end{figure*}

\noindent \textbf{Necessity of Sufficient Ablative Experiments.}
The ablation study is recognized as the primary means of evaluating the effectiveness of individual components within a specific method. 
However, we assume that the conclusions drawn from ablative experiments merely on CASIA-B~\cite{yu2006framework} cannot easily extend to real scenarios. 
The primary reasons are two-fold: 
a)~
CASIA-B contains only 50 subjects for testing. Such few subjects for testing make the results vulnerable to noise.
b)~
The silhouettes from a good person segmentation model are different from those from the simple background subtraction algorithm of CASIA-B.
In principle, performing a comprehensive ablation study across diverse large-scale datasets can yield a robust configuration of hyperparameters.

\label{sec:base}
\noindent \textbf{Necessity of a Strong Backbone.}
The quality of a model largely hinges on the capability of its backbone. If the backbone is not strong enough, it may inflate the effectiveness of additional modules. The evolution of CNN architectures has progressed from shallow to deep with emerging excellent backbone networks like AlexNet~\cite{krizhevsky2012imagenet}, VGG-16~\cite{simonyan2015very}, and ResNet~\cite{he2016deep}. However, previous works~\cite{chao2019gaitset, huang2021context,zheng2022gait3d} in gait recognition have predominantly relied on shallow convolutional neural networks, which are composed of just several convolution layers. As some large-scale real-world datasets~\cite{zhu2021gait,zheng2022gait3d} become accessible, the necessity of a robust and powerful gait backbone becomes apparent and essential.

\section{Baselines for Gait Recognition}
\label{sec:gait_model}
Building on the insights outlined previously, this section introduces a series of baseline models that reveal several  practical principles for in-the-wild gait recognition. 
Our investigation encompasses appearance-based, model-based, and multi-modal methodologies, contributing to the comprehensiveness of this benchmark study. 
To ensure broader applicability, we prioritize the development of structurally simple yet experimentally powerful and empirically robust architectures to lay the foundations for follow-ups. 
To this end, we undertake three-fold efforts: 
\begin{itemize}
    \item \textbf{A General Pipeline}. 
    To ensure the baseline models' typicality and universality, a widely recognized pipeline shown in Fig.~\ref{fig:pipeline} (a) has been adopted. 
    \item \textbf{Two Basic Blocks}. 
    To make it simple and scaleable, only two blocks are employed for the network backbone. As illustrated in Fig.~\ref{fig:pipeline} (b), they are the 3D residual unit and its pseudo counterpart~\cite{hara2017learning, qiu2017learning}. 
    \item \textbf{Three Strong Baselines}. 
    As depicted in Fig.~\ref{fig:pipeline} (c), the three baselines, DeepGaitV2, SkeletonGait, and SkeletonGait++, share an identical pipeline. They take the silhouette images, skeletal coordinates, and their combination as the input, respectively representing the appearance-based, model-based, and multi-modal baseline model.
    Notably, we align the data format of silhouettes and skeletons by drawing the latter's coordinates into an imagery heatmap. It enables a flexible comparison between these two distinct gait modalities. Details are provided in Sec.~\ref{sec:skeletongait}.
\end{itemize}

The subsequent content begins by introducing the proposed pipeline,
followed by the introductions to the three baselines, DeepGaitV2, SkeletonGait and SkeletonGait++.

\subsection{The Pipeline}
As shown in Fig.~\ref{fig:pipeline} (a), the backbone transforms each input frame into a 3D feature map with the height, width, and channel dimensions.
Then, a temporal pooling (TP) module aggregates the obtained feature map sequence by max pooling along the temporal dimension, it outputs a global understanding of the data.
Next, the obtained feature map is horizontally divided into several parts, and each part is pooled into a feature vector, according to the popular horizontal pooling (HP) in~\cite{fu2019horizontal}.
As a result, we can get several feature vectors and use separate fully connected layers (FCs) to map them into the metric space, \textit{i.e.}, generating $f$.
Since these vectors will be processed independently for each part, our formulation loosely treats $f$ as a single feature vector for brevity.

Given a mini-batch $\mathbb{B}$, we can get a collection of positive and negative pairs over this batch, namely $\mathbb{S}^+$ and $\mathbb{S}^-$. 
For each sample pair $(f^{+}_{i}, f^{-}_{i})$ where $i\in\mathbb{S}^+ \cup \mathbb{S}^-$, we utilize the Euclidean norm to measure their distance, \textit{i.e.}, $d_i=\left\| f^{+}_{i} - f^{-}_{i} \right\|$. 
Then the triplet loss is employed to drive identity learning: 
\begin{equation}
        L_{triplet} = \sum_{i\in \mathbb{S}^+} \sum_{j\in \mathbb{S}^-} \left\lfloor d_i - d_j + m \right\rfloor_+ 
\label{equ:triplet}
\end{equation}
where $\left\lfloor \cdot \right\rfloor_+$ denotes the ReLU function~\cite{6302935} and $m$ is the loss margin~\cite{Chen_2017_CVPR} with a default value of 0.2. In practice, $L_{triplet}$ is further reduced by the number of non-zero terms over $\left\lfloor \cdot \right\rfloor_+$. 

Next, we use BNNecks~\cite{luo2019bag} to optimize the identity distribution.
For each $i$-th sample within batch $\mathbb{B}$, 
we let $p^i_{\hat{y}}$ present its probability belonging to class $\hat{y}\in \{ 1, 2, ..., \text{Y} \}$, where $\text{Y}$ is the class number. 
Given the ground-truth label $y^i$, the cross-entropy loss can be formulated as: 
\begin{equation}
    \begin{aligned}
        L_{ce} &= - \frac{1}{\mathcal{R}(\mathbb{B})} \sum_{i\in\mathbb{B}} \frac{1}{\text{Y}} \sum_{\hat{y}\in\{1, 2, ..., \text{Y}\}} \text{log} (p^i_{y})
        \\
        p_{y}^i &= p_{\hat{y}}^i \quad \text{if} \quad \hat{y}=y^i \quad \text{otherwise} \quad 1 - p_{\hat{y}}^i
   \end{aligned}
\label{equ:ce}
\end{equation}
where $\mathcal{R}(\cdot)$ measures the sample number within batch $\mathbb{B}$. 
In practice, a soft classification trick~\cite{hou2020gait} is further employed to encourage the model to be less confident on the training set.


\begin{table}[t]
\centering
\caption{
Architectures of DeepGaitV2 series. 
}
\renewcommand{\arraystretch}{1.0}
\begin{threeparttable}
\begin{tabular}{c|c|ccccc}
\toprule
\multirow{3}{*}{Layer} & \multirow{3}{*}{\begin{tabular}[c]{@{}c@{}}Output\\ Size\end{tabular}} & \multicolumn{5}{c}{DeepGaitV2}                 \\ \cline{3-7} 
& & \multicolumn{1}{c|}{\multirow{2}{*}{Block}} & \multicolumn{4}{c}{Network Depth $D$ } \\ \cline{4-7} 
& & \multicolumn{1}{c|}{}                      & 10   & 14   & 22   & 30   \\ \hline \hline
Conv 0                  & ($T$,$C$,64,44)                                                                       & \multicolumn{5}{c}{$3\times3$, stride 1}                              \\ \midrule
Stage 1                 & ($T$,$C$,64,44) & \multicolumn{1}{c|}{$\begin{bmatrix}
3\times3, C\\ 
3\times3, C
\end{bmatrix}$}                      & $\times1$     & $\times1$      & $\times1$     & $\times1$     \\ 
Stage 2                 & ($T$,$2C$,32,22) & \multicolumn{1}{c|}{$\begin{bmatrix}
3\times3\times3, 2C\\ 
3\times3\times3, 2C
\end{bmatrix}$}                      & $\times1$     & $\times2$     & $\times4$     & $\times4$     \\
Stage 3                 & ($T$,$4C$,16,11) & \multicolumn{1}{c|}{$\begin{bmatrix}
3\times3\times3, 4C\\ 
3\times3\times3, 4C
\end{bmatrix}$}                      & $\times1$     & $\times2$     & $\times4$     & $\times8$     \\
Stage 4                 & ($T$,$8C$,16,11) & \multicolumn{1}{c|}{$\begin{bmatrix}
3\times3\times3, 8C\\ 
3\times3\times3, 8C
\end{bmatrix}$}                      & $\times1$     & $\times1$     & $\times1$     & $\times1$     \\ \midrule
TP                     & (1,$8C$,16,11) & \multicolumn{5}{c}{Temporal Pooling}              \\
HP                     & (1,$8C$,16,1) & \multicolumn{5}{c}{Horizontal Pooling}            \\
Head                   & (1,$8C$,16,1) & \multicolumn{5}{c}{Separate FCs and BNNecks}      \\ 
\bottomrule
\end{tabular}
\textit{
Note: Down-sampling is performed by Stage 2 and 3 with a stride of 2. 
$T$ and $C$ respectively denote the sequence length and arbitrary channel number.}
\end{threeparttable}
\label{tab:deepgaitv2}
\end{table}

\subsection{Appearance-based DeepGaitV2}
\label{sec:deepgaitv2}
Gait data is usually in a simple format, such as the silhouette with a size of $64\times44$. It creates the impression that gait recognition is an \textit{easy} task that shallow networks can handle. 
In the context of developing baseline models, we agree that the gait recognition simulated by constrained gait datasets, such as the CASIA-B~\cite{yu2006framework}, OU-MVLP~\cite{takemura2018multi}, and the latest SUSTech1K~\cite{Shen_2023_CVPR} and CCPG~\cite{Li_2023_CVPR} datasets, can be considered as relatively simple tasks that shallow gait models can deal with.
The reason is that they contain limited covariates manipulated in data collection. The models can circumvent human-made challenges by focusing on evidently unchanged cues rather than subtle walking patterns~\cite{Li_2023_CVPR}. 
Therefore, we assume that deep gait models easily overfit on in-the-lab datasets.
To validate this hypothesis, we build the DeepGaitV2 series. 

\begin{figure*}[tb]
\centering
\includegraphics[width=\linewidth]{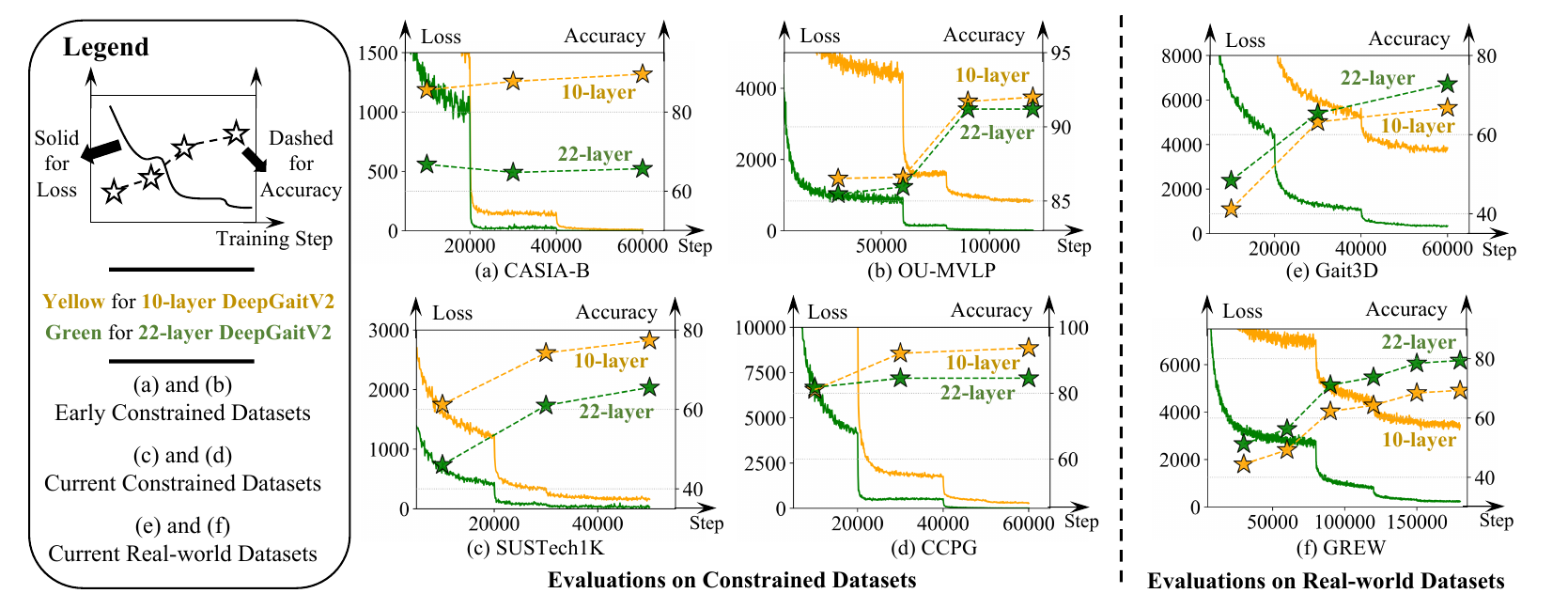}
\caption{
The DeepGaitV2 series meets the over-fitting cases on (a) CASIA-B and (b) OU-MVLP, with the network depth increasing. 
The loss number presents the count of triplets that cause a non-zero loss in the training batch, directly reflecting the network's convergence state.
}
\label{fig:overfitting} 
\end{figure*}

As shown in Table~\ref{tab:deepgaitv2}, the backbone of DeepGaitV2 is initialized as a stack of five layers, \textit{i.e.}, the initial Conv 0 and the following Stage 1 to 4. 
The number of residual blocks within each stage, denoted as $D$, can scale the network depth.
For instance, if $D$ is set to [1,1,1,1], the network depth should be 2$\times$($1$+$1$+$1$+$1$)+2=10, where the last +2 implies the initial Conv 0 and final head layers.
As a result, Table~\ref{tab:deepgaitv2} presents a serial of DeepGaitV2 with a depth ranging from 10 to 30. 
Note we instantiate Stage 1 as pure 2D residual block~\cite{he2016deep} for computational efficiency. 

\begin{table}[t]
\centering
\caption{
The Rank-1 Accuracies by DeepGaitV2 with Various Depths.
}
\renewcommand{\arraystretch}{1.0}
\begin{threeparttable}
\setlength{\tabcolsep}{1pt}
\begin{tabular}{c|ccc|cc|cc}
\toprule
\multirow{2}{*}{Method}     & \multirow{2}{*}{Depth} & \multirow{2}{*}{\textcolor{blue}{Param.}} & \multirow{2}{*}{\textcolor{blue}{GFLOPs}} & \multicolumn{2}{c|}{In-the-lab} & \multicolumn{2}{c}{In-the-wild} \\ \cline{5-8}
                            &   & &                     & CASIA-B        & OU-MVLP        & Gait3D           & GREW         \\ \hline \hline
GaitSet~\cite{chao2019gaitset}                     & $<$10 & \textcolor{blue}{2.3} & \textcolor{blue}{12.9}                & 87.1           & 87.1           & 36.7             & 48.4         \\
GaitPart~\cite{fan2020gaitpart}                    & $<$10 & \textcolor{blue}{3.8} & \textcolor{blue}{7.9}               & 88.5           & 88.7           & 28.2             & 47.6         \\
GaitGL~\cite{lin2021gait}                      & $<$10     & \textcolor{blue}{7.0} & \textcolor{blue}{58.5}            & \textbf{91.9}           & \textbf{92.0}           & 29.7             & 47.3         \\
GaitBase~\cite{Fan_2023_CVPR}                    & 10      & \textcolor{blue}{4.9} & \textcolor{blue}{35.5}              & 89.4           & 90.8           & 60.1             & 60.1         \\ \midrule
\multirow{4}{*}{\begin{tabular}[c]{@{}c@{}}DeepGaitV2 \\ (Shallow \\ to Deep) \end{tabular}} 
                            & 10                  & \textcolor{blue}{14.2} & \textcolor{blue}{91.5}  & 89.6           & \textbf{92.0}           & 66.8             & 69.3         \\
                            & 14                  & \textcolor{blue}{18.6} & \textcolor{blue}{129.0}  & 85.5$\downarrow$           & \textbf{92.0}           & 70.8$\uparrow$             & 75.7$\uparrow$         \\
                            & 22                  & \textcolor{blue}{27.5} & \textcolor{blue}{203.7}  & 75.3$\downarrow$           & 91.9$\downarrow$            & \textbf{72.8}$\uparrow$             & 79.4$\uparrow$         \\
                            & 30                  & \textcolor{blue}{41.6} & \textcolor{blue}{278.5}  & 65.7$\downarrow$           & 91.2$\downarrow$            & 71.7$\downarrow$             & \textbf{79.5}$\uparrow$         \\ 
\bottomrule
\end{tabular}
\textit{Note: the rank-1 accuracy on CASIA-B is the average over its three conditions. Param. refers to model size (M).}
\end{threeparttable}
\label{tab:overfitting}
\end{table}

With deepening the backbone, as illustrated in Table~\ref{tab:overfitting}, there exists a noticeable distinction in performance trends between in-the-lab and in-the-wild evaluations. 
Specifically, applications on Gait3D~\cite{zheng2022gait3d} and GREW~\cite{zhu2021gait} exhibit substantial benefits from properly deep networks, while opposite outcomes are observed on CASIA-B~\cite{yu2006framework} and OU-MVLP~\cite{takemura2018multi}.
The analysis shown in Fig.~\ref{fig:overfitting} (a) and (b) uncover the over-fitting issues on CASIA-B and OU-MVLP, \textit{i.e.}, the 22-layer DeepGaitV2 converges better but performs worse. 

Moreover, we find similar over-fitting phenomena on the latest constrained gait datasets,  SUSTech1K~\cite{Shen_2023_CVPR} and CCPG~\cite{Li_2023_CVPR}.
Despite the introduction of new human-controlled challenges such as poor illumination, complex occlusion, and clothing changes, as shown in Fig.~\ref{fig:overfitting} (c) and (d), the 22-layer DeepGaitV2 continues to exhibit the over-fitting behavior. 

Combining the above observations, this paper advocates for the adoption of deep models as the fundamental architecture for analyzing in-the-wild gait datasets in future research endeavors. 
According to Table~\ref{tab:overfitting}, we utilize the 10-layer and 22-layer DeepGaitV2 models as baseline models for in-the-lab and in-the-wild evaluations, respectively.
Beyond deepening DeepGaitV2, we also develop a light version for better accuracy-speed balance. 
Specifically, we employ pseudo 3D block~\cite{qiu2017learning} shown in Fig.~\ref{fig:pipeline} (b) to reconstruct DeepGaitV2 shown in Table~\ref{tab:deepgaitv2}. 
This slight change can save about 59\% training weights and 57\% computation costs\footnote{
\# param: 27.5 \textit{vs.} 11.1 MB, 
and FLOPs: 6.8 \textit{vs.} 2.9 GFLOPs per frame. Here we only consider the backbone (the same below unless otherwise stated).} with remaining competitive performances on both Gait3D~\cite{zheng2022gait3d} (+1.6\%) and GREW~\cite{zhu2021gait} (-1.7\%).
In the following, we use the pseudo-3D DeepGaitV2 as the default unless explicitly stated. 

\begin{figure*}[t]
\centering
\includegraphics[height=3.5cm]{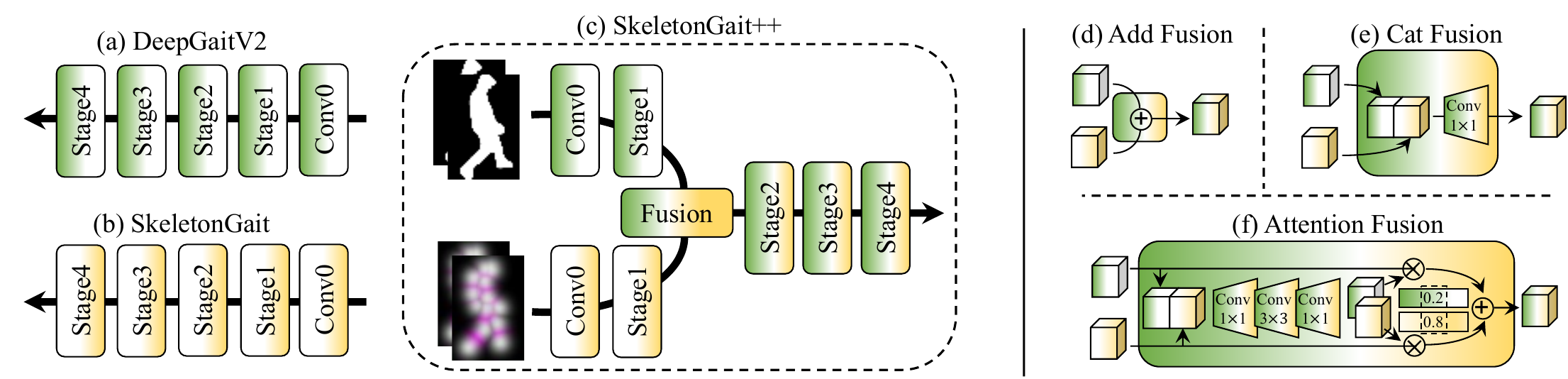}
\caption{
The network architectures of (a) DeepGaitV2, (b) SkeletonGait and (c) SkeletonGait++. The \textit{head} part is ignored for brevity. The different fusions in SkeletonGait++ are (d) add fusion, (e) cat fusion and (f) attention fusion.
}
\label{fig:skeletongait}
\end{figure*}

\subsection{Model-based SkeletonGait}
\label{sec:skeletongait}
Previous skeleton-based methods typically model the coordinates of body joints as non-grid graphs using graph neural networks~\cite{liao2020model,teepe2021gaitgraph,li2020end}. 
In this benchmark study, we introduce a novel skeleton-based gait representation called the skeleton map, drawing inspirations from related works in~\cite{duan2022revisiting, liu2018recognizing, liao2022posemapgait}. 
As illustrated in Fig.~\ref{fig:pipeline} (c), our skeleton map represents the coordinates of joints as a heatmap, \textit{i.e.}, transforming the skeleton into a silhouette-like image without exact body shapes. 
This manner offers three significant advantages: 
\begin{itemize}
    \item Aligning the skeleton and silhouette data formats can enable a seamless extension of DeepGaitV2 to skeleton-based methodology. It eliminates the need for additional efforts to develop graph-specific gait models. This alignment ensures consistency in the network architecture between silhouette-based and skeleton-based gait recognition. So it makes further comparisons intuitive and fair.
    \item The skeleton map, resembling a silhouette without body shapes, allows for an intuitive controlled experiment to observe the distinct roles of body structure \textit{vs.} shape features in gait recognition. 
    \item Serving as an imagery input, the skeleton map can be seamlessly integrated into image-based gait models, particularly at the bottom convolution stages. It can facilitate the further exploration of multi-modal baseline models. 
\end{itemize}

As a result, SkeletonGait is developed by replacing the input of DeepGaitV2 from silhouettes to skeleton maps, as shown in Fig.~\ref{fig:skeletongait} (b). 
The only difference is the number of input channels, as the silhouettes are single-channel while the skeleton maps are double-channel.
Specifically, the skeleton map can be generated by the following steps.

\begin{figure}[t]
\centering
\includegraphics[height=3.2cm]{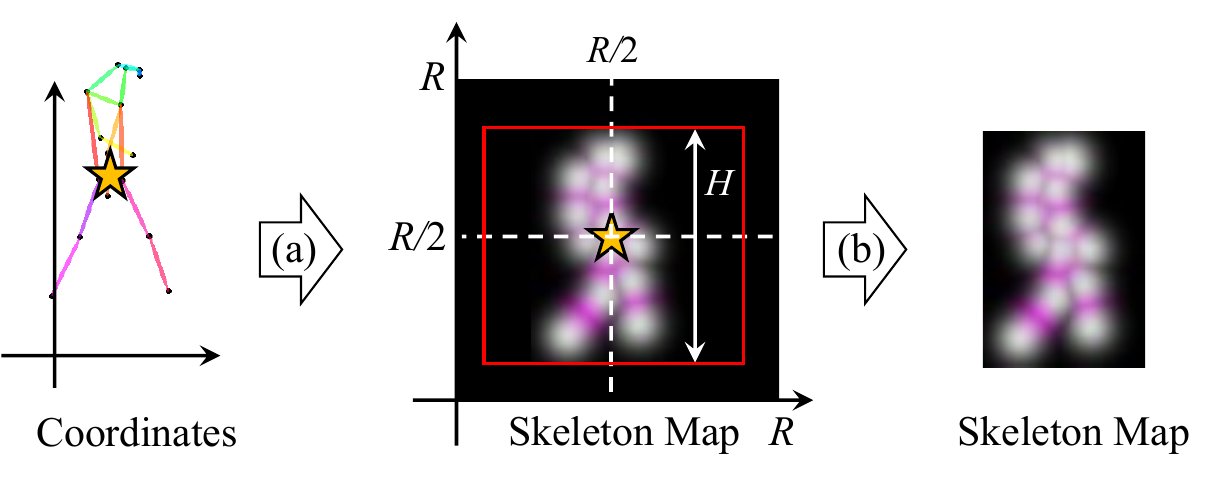}
\caption{
The pipeline of skeleton map generation. (a) Center-normalization, scale-normalization, and skeleton rendering. (b) Subject-centered cropping.
}
\label{fig:generating_skeleton_map}
\end{figure}

Given the coordinates of human joints $(x_k, y_k, c_k)$, where $(x_k, y_k)$ and $c_k$ respectively present the location and confidence score of the $k$-th joint with $k\in\{1, ..., K\}$ ($K$ denotes the number of body joints). 
Firstly, considering the absolute coordinates of joints relative to the original image contain much gait-unrelated information like the walking trajectory and filming distance, we introduce a center-normalization and a scale-normalization to align the coordinates: 
\begin{equation}
    \begin{aligned}
        x_k &= x_k - x_{\text{core}} + R/2 \\ 
        y_k &= y_k - y_{\text{core}} + R/2 \\ 
        x_k &= \frac{x_k - y_{\text{min}}}{y_{\text{max}} - y_{\text{min}}} \times H\\ 
        y_k &= \frac{y_k - y_{\text{min}}}{y_{\text{max}} - y_{\text{min}}} \times H
    \end{aligned}
\label{equ:normalization}
\end{equation}
where $(x_{\text{core}}, y_{\text{core}})=(\frac{x_{11} + x_{12}}{2}, \frac{y_{11} + y_{12}}{2})$ presents the center point of two hips (11-th and 12-th human joints, their center can be regarded as the barycenter of the human body),
and $(y_{\text{max}}, y_{\text{min}})$ denotes the maximum and minimum heights of human joints. 
In this way, we move the barycenter of the human body to $(R/2, R/2)$ and normalize the body height to $H$, as shown in Fig.~\ref{fig:generating_skeleton_map} (a).

Typically, the height of the human body is expected to exceed its width.
As a result, the normalized coordinates of human joints, as defined in Eq.~(\ref{equ:normalization}), should fall within the range of $H \times H$. 
But in practice, the pose estimator is imperfect and may produce some outlier joints outside the $H\times H$ scope. 
To address these out-of-range cases, the resolution of the skeleton map, denoted as $R$, should be larger than $H$. Then all the coordinates will be inside the map. 
In our experiments, we set $R = 2H$ for all datasets. 

As illustrated in Fig.~\ref{fig:generating_skeleton_map} (a), the skeleton map is initialized as a blank image with a size of $R \times R$.
Then it is drawn based on the normalized coordinates of human joints.
Inspired by~\cite{duan2022revisiting}, the joint map $\textit{\textbf{J}}$ is generated by composing $K$ Gaussian maps, where each Gaussian map is centered at a specific joint position and contributes to all the $R \times R$ pixels: 
\begin{equation}
    \textit{\textbf{J}}_{(i, j)}=\sum_{k}^{K}e^{-\frac{(i - x_k)^2 + (j - y_k)^2}{2\sigma^2}} \times c_k
\label{equ:joints}
\end{equation}
where $\textit{\textbf{J}}_{(i, j)}$ presents the value of a certain point from $\{(i, j)|i,j\in \{1, ..., R\}\}$, and $\sigma$ is a hyper-parameter controlling the variance of Gaussian maps. 

Similarly, a limb map $\textit{\textbf{L}}$ can also be created: 
\begin{equation}
    \textit{\textbf{L}}_{(i, j)}=\sum_{n}^{N}e^{-\frac{\mathcal{D}((i,j), \mathcal{S}[n^-, n^+])^2}{2\sigma^2}} \times \text{min}(c_{n^-}, c_{n^+})
\label{equ:limbs}
\end{equation}
where $\mathcal{S}[n^-, n^+]$ presents the $n$-th limb determined by $n^-$-th and $n^+$-th joints with $n^-, n^+\in \{1, ..., K\}$. 
The function $\mathcal{D}((i,j), \mathcal{S}[n^-, n^+])$ measures the Euclidean distance from the point $(i, j)$ to $n$-th limb, where $n\in\{1, ..., N\}$ and $N$ denotes the count of limbs. 

Next, a skeleton map can be obtained by stacking $\textit{\textbf{J}}$ and $\textit{\textbf{L}}$, and thus has a size of $2\times R \times R$.
Notably, for the convenience of visualization, we repeat the last channel of all the skeleton maps shown in this paper to display the visual three-channel images with the size of $3\times R \times R$. 

\begin{figure}[t]
\centering
\includegraphics[height=6.5cm]{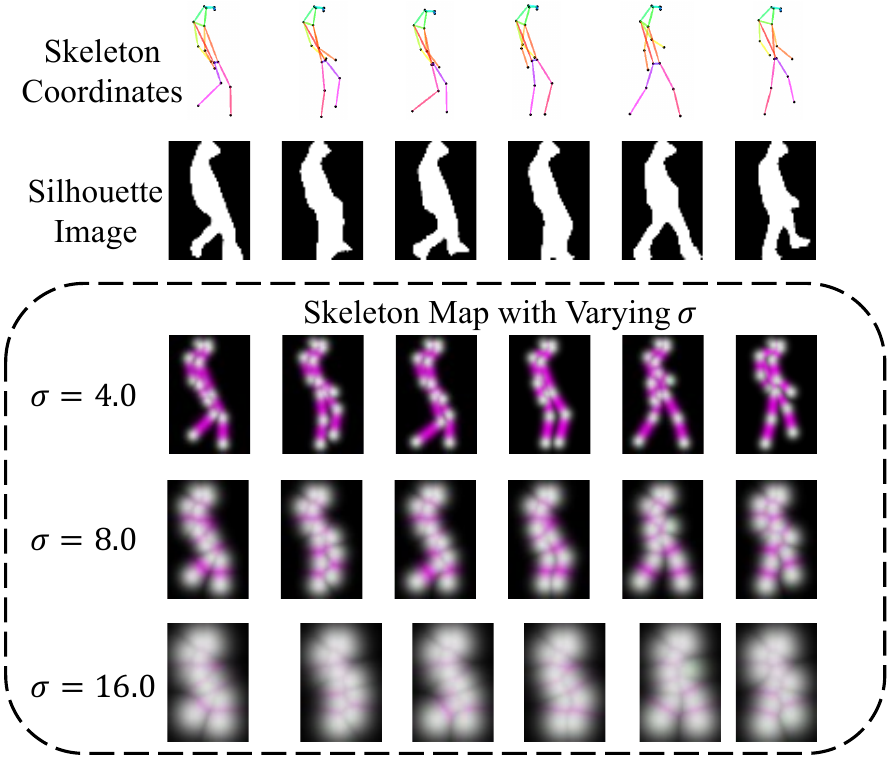}
\caption{
Skeleton coordinates \textit{vs.} silhouette images \textit{vs.} skeleton maps.
}
\label{fig:skeletonmaps}
\end{figure}

As shown in Fig.~\ref{fig:generating_skeleton_map} (b), subject-centered cropping is employed to remove the blank regions. It can reduce the redundancy in skeleton maps. 
In our experiments, the vertical range is determined by the minimum and maximum heights of pixels that possess non-zero values. 
Meanwhile, the horizontal cropping range spans from $\frac{R-H}{2}$ to $\frac{R+H}{2}$. 
In this way, extraneous areas outside the desired gait region can be removed. It ensures a concise and compact skeleton map.
Lastly, to align with the input size required by downstream gait models, the cropped skeleton maps are resized to $2\times 64 \times 64$ and further cropped by the widely-used double-side cutting strategy. 

Fig.~\ref{fig:skeletonmaps} presents some examples of the used skeleton maps with varying $\sigma$.
It can be found that a smaller $\sigma$ produces a visually thinner skeleton map, whereas excessively large $\sigma$ may lead to visual ambiguity. 

Compared with other skeleton-to-image approaches~\cite{duan2022revisiting, liu2018recognizing, liao2022posemapgait}, our skeleton map introduces the following gait-oriented enhancements: 
\begin{itemize}
    \item 
    \textbf{Cleanness}. The implementation of center-normalization effectively eliminates identity-unrelated noise present in raw skeleton coordinates, \textit{i.e.}, the walking trajectory and camera distance information.
    \item 
    \textbf{Discriminability}. Directly resizing the obtained images with varying sizes into a predetermined fixed size, which is used in some previous methods, will inevitably result in the loss of body ratio information. Conversely, the scale-normalization and subject-centered cropping introduced in this paper ensure that the skeleton map preserves the authenticity of the length and ratio of human limbs.
    \item 
    \textbf{Compactness}. All the joints and limbs are drawn within a single map, optimizing the efficiency of the modeling process, as opposed to a stack of separate maps.
\end{itemize}

\subsection{Multi-modal SkeletonGait++}
To integrate the superiority of silhouettes and skeleton maps, as shown in Fig.~\ref{fig:skeletongait} (c), 
SkeletonGait++ provides a fusion-based two-stream architecture with a silhouette branch and a skeleton one. 
These two branches respectively share the same network architectures with DeepGaitV2 and SkeletonGait at their early stages, such as Conv 0 and Stage 1.
Then, a fusion module is to aggregate the two feature sequences from the two branches in a frame-by-frame manner. 
Three kinds of fusion mechanisms are considered and evaluated in the experiments. 
\begin{itemize}
    \item \textbf{Add Fusion.} The feature maps from the silhouette and skeleton branches are combined using an element-wise addition operation, as demonstrated in Fig.~\ref{fig:skeletongait}~(d).
    \item \textbf{Concatenate Fusion.} The feature maps from the silhouette and skeleton branch are first concatenated along the channel dimension, and then transformed by a plain $1\times1$ convolution layer, as demonstrated in Fig.~\ref{fig:skeletongait}~(e). 
    \item \textbf{Attention Fusion.} The feature maps from the silhouette and skeleton branch are first concatenated along the channel dimension, and then transformed by a small network to form a cross-branch understanding. 
    Here the small network is composed of a squeezing $1\times1$, a plain $3\times3$, and an expansion $1\times1$ convolution layer. 
    As shown in Fig.~\ref{fig:skeletongait}~(f), a softmax layer is employed to assign element-wise attention scores respectively for the silhouette and skeleton branches. 
    Lastly, an element-wise weighted-sum operation is used to generate the output.
\end{itemize}

In this work, SkeletonGait++ considers two fusion locations. 
The first is a low-level counterpart as illustrated in Fig.~\ref{fig:skeletongait}~(c), where the fusion module works before Stage 2. 
Another high-level fusion aggregates the features before Stage 4, with additional Stage 2 and 3 respectively being inserted into the silhouette and skeleton branches. 

\section{More Experiments}
\label{sec:exp}
For a convincing elaboration, the key statistics of employed datasets in the experiments are presented in Table~\ref{tab:gait_datasets}. The experimental designs have been briefly described in the previous section along with the description of DeepGaitV2. In this section, the implementation details are provided for completeness, and then the comparison and ablation studies are presented.

\begin{table}[tb]
\centering
\caption{Some Implementation details.
}
\resizebox{0.46\textwidth}{!}{
\begin{threeparttable}
\begin{tabular}{cccc}
\toprule
 DataSet & Batch Size & Milestones & Total Steps \\ \hline \hline
 CASIA-B~\cite{yu2006framework}             & (8, 16)                    & (20k, 40k, 50k)               & 60k                    \\ 
 OU-MVLP~\cite{takemura2018multi}              & (32, 8)                    & (60k, 80k, 100k)              & 120k                   \\ 
 CCPG~\cite{Li_2023_CVPR}                & (8, 16)                    & (20k, 40K, 50k)               & 60k                    \\ 
 SUSTech1K~\cite{Shen_2023_CVPR}           & (8,  8)                    & (20k, 30k, 40k)               & 50k                    \\
 Gait3D~\cite{zheng2022gait3d}              & (32, 4)                    & (20k, 40K, 50k)               & 60k                    \\ 
 GREW~\cite{zhu2021gait}                & (32, 4)                    & (80k, 120k, 150k)             & 180k                   \\
 \bottomrule 
\end{tabular}
\textit{Note: batch size $(8, 16)$ denotes $8$ subjects and $16$ samples per subject.} \\
\end{threeparttable}
}
\label{tab:implementations}
\end{table}

\noindent \textbf{Implementation Details.} 
Table~\ref{tab:implementations} lists the main hyper-parameters of our experiments. Unless otherwise specified, the settings are as follows. 
\begin{itemize}
\item The silhouettes are aligned by the normalization strategy in~\cite{takemura2018multi} and resized to $64\times44$. The spatial augmentation strategy outlined in~\cite{Fan_2023_CVPR} is adopted.
\item Different datasets may have different skeleton  formats, such as COCO~18 for OU-MVLP and BODY~25 for CCPG. 
To enhance flexibility, in our experiments all formats are standardized to COCO~17.
\item At the test phase, a whole gait sequence is fed into the model directly whatever how many frames it has. But for training the data sampler collects a fixed-length clip of 30 frames as the input. 
\item The SGD optimizers are with an initial learning rate of 0.1 and a weight decay of 0.0005. 
The learning rate will be reduced by 0.1 at each milestone in Table~\ref{tab:implementations}. 
\item For DeepGaitV2, the channel $C$ in Table~\ref{tab:deepgaitv2} is set to a default value of 64. 
For SkeletonGait, the variance $\sigma$  in Eq.~(\ref{equ:joints}) and (\ref{equ:limbs}) is set to a default value of 8.0.
The fusion strategy in SkeletonGait++ is instantiated as the low-level attention fusion as depicted in Fig.~\ref{fig:skeletongait} (c) and (f). 
\item 
All the recognition metrics mentioned below are estimated according to the evaluation protocol provided by the dataset creators. For instance, the Rank-$k$ (R-$k$) accuracy represents the identification rate compared with the top $k$ ranking samples in the gallery set. Specifically for the Gait3D dataset, the mean Average Precision (mAP) and mean Inverse Negative Penalty (mINP)~\cite{ye2021deep}, which consider the recall of multiple instances and hard samples, have also been included in the comparison.

\item All the implementations are by OpenGait available at \url{https://github.com/ShiqiYu/OpenGait}. 
\end{itemize}

\begin{table*}[tb]
\centering
\caption{
Recognition results on the OU-MVLP~\cite{takemura2018multi}, Gait3D~\cite{zheng2022gait3d}, and GREW~\cite{zhu2021gait} datasets. 
}
\resizebox{2.0\columnwidth}{!}{
\begin{threeparttable}
\begin{tabular}{c|cc|ccccccccc}
\toprule
\multirow{3}{*}{Category}      & \multirow{3}{*}{Method} & \multirow{3}{*}{\textcolor{blue}{Year}} & \multicolumn{9}{c}{Testing Datasets}                                                                                     \\ \cline{4-12} 
                            &                         &                         & \multicolumn{1}{c|}{OU-MVLP~\cite{takemura2018multi}} & \multicolumn{4}{c|}{Gait3D~\cite{zheng2022gait3d}} & \multicolumn{4}{c}{GREW~\cite{zhu2021gait}}                                     \\ \cline{4-12} 
                            &                         &                         & \multicolumn{1}{c|}{Rank-1}  & Rank-1 & Rank-5  & mAP  & \multicolumn{1}{c|}{mINP} & Rank-1 & Rank-5 & Rank-10 & Rank-20  \\ \hline \hline 
\multirow{9}{*}{\begin{tabular}[c]{@{}c@{}} Silhouette-  \\ based \end{tabular}} 
& GaitSet~\cite{chao2019gaitset}                 & \textcolor{blue}{2019}                & \multicolumn{1}{c|}{87.1}    & 36.7   & 58.3    & 30.0 & \multicolumn{1}{c|}{17.3} & 46.3   & 63.6   & 70.3    & -        \\
& GaitPart~\cite{fan2020gaitpart}                & \textcolor{blue}{2020}                & \multicolumn{1}{c|}{88.5}    & 28.2   & 47.6    & 21.6 & \multicolumn{1}{c|}{12.4} & 44.0   & 60.7   & 67.3    & -        \\
& GaitGL~\cite{lin2021gait}                  & \textcolor{blue}{2021}                & \multicolumn{1}{c|}{89.7}    & 29.7   & 48.5    & 22.3 & \multicolumn{1}{c|}{13.6} & 47.3   & \multicolumn{3}{c}{-}                           \\
& DANet~\cite{ma2023dynamic}                   & \textcolor{blue}{2023}                & \multicolumn{1}{c|}{90.7}    & 48.0     & 69.7 & - & \multicolumn{1}{c|}{-} & \multicolumn{4}{c}{-}    \\
& GaitBase~\cite{Fan_2023_CVPR}                & \textcolor{blue}{2023}                & \multicolumn{1}{c|}{90.8}    & 64.6     & \multicolumn{3}{c|}{-} & 60.1 & \multicolumn{3}{c}{-}    \\
& GaitSSB~\cite{fan2022learning}                & \textcolor{blue}{2023}                & \multicolumn{1}{c|}{91.8}    & 63.6     & \multicolumn{3}{c|}{-} & 61.7 & \multicolumn{3}{c}{-}    \\
& QAGait~\cite{wang2024qagait}                  & \textcolor{blue}{2024}                & \multicolumn{1}{c|}{-}    & 67.0     & 81.5 & 56.5 & \multicolumn{1}{c|}{-} & 59.1 & 74.0 & 79.2 & 83.2  \\
                            \cmidrule{2-12}
& DeepGaitV2          & Ours               & \multicolumn{1}{c|}{\textbf{91.9}}    & \textbf{74.4}   & \textbf{88.0}    & \textbf{65.8} & \multicolumn{1}{c|}{\textbf{39.2}}  & \textbf{77.7}   & \textbf{88.9}   & \textbf{91.8}    & \textbf{93.0}          \\ \midrule
\multirow{5}{*}{\begin{tabular}[c]{@{}c@{}} Skeleton-  \\ based \end{tabular}}   & GaitGraph2~\cite{teepe2021gaitgraph}              & \textcolor{blue}{2022}               & \multicolumn{1}{c|}{62.1}    & 11.1   & \multicolumn{3}{c|}{-} & 33.5   & \multicolumn{3}{c}{-}                           \\
                            & Gait-TR~\cite{zhang2022spatial}                  & \textcolor{blue}{2023}               & \multicolumn{1}{c|}{56.2}    & 6.6    & \multicolumn{3}{c|}{-} & 54.5   & \multicolumn{3}{c}{-}                           \\
                            & GPGait~\cite{Fu_2023_ICCV}                  & \textcolor{blue}{2023}                & \multicolumn{1}{c|}{60.5}    & 22.5   & \multicolumn{3}{c|}{-} & 53.6   & \multicolumn{3}{c}{-}                           \\ \cmidrule{2-12}
& SkeletonGait                & Ours                    & \multicolumn{1}{c|}{\underline{67.4}}    & \underline{38.1}   & \underline{56.7}    & \underline{28.9} & \multicolumn{1}{c|}{\underline{16.1}} & \underline{77.4}   & \underline{87.9}   & \underline{91.0}    & \underline{93.2}       \\ \midrule
\multirow{6}{*}{\begin{tabular}[c]{@{}c@{}}Multi-modal\end{tabular}} 
& SMPLGait~\cite{zheng2022gait3d}                & \textcolor{blue}{2022}                & \multicolumn{1}{c|}{-}       & 46.3   & 64.5    & 37.2 & \multicolumn{1}{c|}{22.2} & \multicolumn{4}{c}{-}                                    \\ 
& GaitRef~\cite{zhu2023gaitref}                & \textcolor{blue}{2023}                & \multicolumn{1}{c|}{90.2}   & 49.0   & 49.3    & 40.7 & \multicolumn{1}{c|}{25.3} & 53.0   & 67.9   & 73.0 & 77.5   \\ 
& HybridGait~\cite{dong2024hybridgait}                & \textcolor{blue}{2024}                & \multicolumn{1}{c|}{-}       & 53.3   & 72.0    & 43.3 & \multicolumn{1}{c|}{26.7} & \multicolumn{4}{c}{-}                                    \\ 
\cmidrule{2-12}
& SkeletonGait++          & Ours               & \multicolumn{1}{c|}{-}    & \underline{\textbf{77.6}}   & \underline{\textbf{89.4}}    & \underline{\textbf{70.3}} & \multicolumn{1}{c|}{\underline{\textbf{42.6}}} & \underline{\textbf{85.8}}   & \underline{\textbf{92.6}}   & \underline{\textbf{94.3}}    & \underline{\textbf{95.5}}           \\ \bottomrule 
\end{tabular}
\textit{
Note: The highest results are in} \textbf{bold} \textit{for silhouette-based methods and} \underline{underlined} \textit{for skeleton-based methods. 
Regarding multi-modal methods, the highest results are in both} \textbf{\underline{bold and underlined}}. \textit{The same annotations are applied in the following tables. }
\end{threeparttable} 
}
\label{tab:main_results}
\end{table*}

\subsection{Analysis by the Appearance-based DeepGaitV2}
Except for the degradation observed on CASIA-B~\cite{yu2006framework} (Table~\ref{tab:overfitting}), DeepGaitV2 outperforms other silhouette-based methods across all other datasets, as demonstrated in Table~\ref{tab:main_results}, \ref{tab:results_on_SUSTech1K}, and \ref{tab:results_on_CCPG}.
Particularly, DeepGaitV2 exhibits more substantial improvements on real-world large-scale gait datasets, such as Gait3D~\cite{zheng2022gait3d} and GREW~\cite{zhu2021gait},
than on constrained datasets like OU-MVLP~\cite{takemura2018multi}, SUSTech1K~\cite{Shen_2023_CVPR}, and CCPG~\cite{Li_2023_CVPR}. This alignment with the primary objective of our benchmark study underscores its role in advancing gait recognition in real-world applications.

Moreover, Table~\ref{tab:overfitting} illustrates that the performance advantage of DeepGaitV2 primarily arises from its depth. The deeper and the larger, the higher rank-1 accuracy can be gained. 
It is straightforward in deep learning for computer vision but was not well addressed in gait recognition.
The overfitting on some small constrained datasets like CASIA-B can mislead the exploring to a deep model. 
Therefore, this work emphasizes the importance of adopting powerful network architectures as the backbone in gait recognition. 

\subsection{Analysis by the Model-based SkeletonGait}
As shown in Table~\ref{tab:main_results}, \ref{tab:results_on_SUSTech1K}, and \ref{tab:results_on_CCPG}, SkeletonGait outperforms the latest skeleton-based methods by breakthrough improvements in most cases. 
It gains $+5.3\%$, $+15.6\%$, $+22.9\%$, $29.2\%$, and $+17.4\%$ rank-1 accuracy on OU-MVLP~\cite{takemura2018multi}, Gait3D~\cite{zheng2022gait3d}, GREW~\cite{zhu2021gait}, SUSTech1K~\cite{Shen_2023_CVPR} and CCPG~\cite{Li_2023_CVPR}, respectively. 

To exclude the potential positive influence brought by the model size of SkeletonGait, we reduced its channels by half. It made its model size nearly identical to that of GPGait~\cite{Fu_2023_ICCV}, \textit{i.e.}, 2.85M v.s. 2.78M. 
Even so, SkeletonGait can still reach the rank-1 accuracy of 33.2\% on Gait3D~\cite{zheng2022gait3d} and 70.9\% on GREW~\cite{zhu2021gait}, and achieves better results than other skeleton-based methods shown in Table~\ref{tab:main_results}.

To demonstrate the robustness of SkeletonGait to different upstream pose estimators, we conduct experiments using both AlphaPose and OpenPose data provided by OU-MVLP~\cite{takemura2018multi}, resulting in a rank-1 accuracy of 67.4\% and 65.9\%, respectively. 
These two results consistently surpass other skeleton-based methods, revealing the robustness of SkeletonGait. 
Note Table~\ref{tab:main_results} only presents the higher one for brevity. 

\begin{table*}[t]
\centering
\caption{
Evaluation with various conditions on SUSTech1K~\cite{Shen_2023_CVPR}.
}
\setlength{\tabcolsep}{0.5em}
\begin{threeparttable}
\begin{tabular}{c|cc|cccccccc|c|c}
\toprule
\multirow{2}{*}{Input} & \multirow{2}{*}{Method}   & \multirow{2}{*}{\textcolor{blue}{Year}}     & \multicolumn{8}{c|}{Probe Sequence (R-1)}      &  \multicolumn{2}{c}{Overall}                    \\
  & & & Normal    & Bag    & Clothing  & Carrying  & Umbrella  & Uniform  & Occlusion & Night   & R-1    & R-5  \\         \hline \hline
\multirow{6}{*}{\begin{tabular}[c]{@{}c@{}}Silhouette- \\ based\end{tabular}} & GaitSet~\cite{chao2019gaitset}      & \textcolor{blue}{2019}         & 69.1     & 68.2   & 37.4     & 65.0     & 63.1      & 61.0    & 67.2     & 23.0   & 65.0  & 84.8  \\
& GaitPart~\cite{fan2020gaitpart}    & \textcolor{blue}{2019}                                & 62.2     & 62.8   & 33.1     & 59.5     & 57.2      & 54.8    & 57.2     & 21.7   & 59.2  & 80.8  \\
& GaitGL~\cite{lin2021gait}        & \textcolor{blue}{2021}                              & 67.1     & 66.2   & 35.9     & 63.3     & 61.6      & 58.1    & 66.6     & 17.9  &  63.1 & 82.8  \\ 
& GaitBase~\cite{Fan_2023_CVPR}    & \textcolor{blue}{2023}                             & 81.5     & 77.5   & 49.6     & 75.8     & 75.5      & 76.7    & 81.4     & 25.9   & 76.1  & 89.4   \\ \cmidrule{2-13} 
& DeepGaitV2         & Ours                        & \textbf{87.4}     & \textbf{84.1}  & \textbf{53.4}    & \textbf{81.3}     & \textbf{86.1}      & \textbf{84.8}    & \textbf{88.5}     & \textbf{28.8}   & \textbf{82.3}   & \textbf{92.5} \\ \midrule
\multirow{5}{*}{\begin{tabular}[c]{@{}c@{}}Skeleton- \\ based\end{tabular}} 
& GaitGraph2~\cite{teepe2021gaitgraph}    & \textcolor{blue}{2022}                                & 22.2     & 18.2   & 6.8     & 18.6     & 13.4      & 19.2    & 27.3     & 16.4   & 18.6  & 40.2  \\
& Gait-TR~\cite{zhang2022spatial}      & \textcolor{blue}{2023}         & 33.3     & 31.5   & 21.0     & 30.4     & 22.7      & 34.6    & 44.9     & 23.5   & 30.8  & 56.0  \\
& MSGG~\cite{peng2021learning}        & \textcolor{blue}{2023}                              & 67.1     & 66.2   & 35.9     & 63.3     & 61.6      & 58.1    & 66.6     & 17.9  &  33.8 & -  \\ \cmidrule{2-13}
& SkeletonGait    & Ours                             & \underline{67.9}     & \underline{63.5}   & \underline{36.5}    & \underline{61.6}     & \underline{58.1}     & \underline{67.2}    & \underline{79.1}     & \underline{50.1}   & \underline{63.0}  & \underline{83.5}   \\ \midrule
\multirow{2}{*}{\begin{tabular}[c]{@{}c@{}}Multi- \\ modal \end{tabular}} & BiFusion~\cite{peng2021learning}      & \textcolor{blue}{2023}         & 69.8     & 62.3   & 45.4     & 60.9     & 54.3      & 63.5    & 77.8     & 33.7   & 62.1  & 83.4  \\
& SkeletonGait++    & Ours                                & \underline{\textbf{89.1}}     & \underline{\textbf{87.2}}   & \underline{\textbf{55.3}}     & \underline{\textbf{85.3}}     & \underline{\textbf{87.3}}      & \underline{\textbf{87.6}}    & \underline{\textbf{91.3}}     & 47.9   & \underline{\textbf{85.6}}  & \underline{\textbf{95.0}}  \\
\bottomrule
\end{tabular}
\end{threeparttable}
\label{tab:results_on_SUSTech1K}
\end{table*}

\begin{table*}[t]
\centering
\caption{
Evaluation With Various Conditions on CCPG~\cite{Li_2023_CVPR}.
}
\begin{threeparttable}
\begin{tabular}{c|cc|cccc|c|cccc|c} 
\toprule
\multirow{2}{*}{Input}    & \multirow{2}{*}{Model}                                     & \multirow{2}{*}{\textcolor{blue}{Year}} & \multicolumn{5}{c|}{Gait Evaluation Protocol}                                                               & \multicolumn{5}{c}{ReID Evaluation Protocol}                                                                 \\ 
\cmidrule{4-13}
                          &                                                            &                        & CL              & UP              & DN              & BG              & Mean            & CL              & UP              & DN              & BG              & Mean             \\ \hline \hline
\multirow{6}{*}{\begin{tabular}[c]{@{}c@{}}Silhouette-\\ based \end{tabular}}     & GaitSet~\cite{chao2019gaitset}                   & \textcolor{blue}{2019}               & 60.2          & 65.2          & 65.1          & 68.5          & 64.8          & 77.5          & 85.0          & 82.9          & 87.5          & 83.2           \\
                          & GaitPart~\cite{fan2020gaitpart}           & \textcolor{blue}{2020}                & 64.3          & 67.8          & 68.6          & 71.7          & 68.1          & 79.2          & 85.3          & 86.5          & 88.0          & 84.8           \\
                          & AUG-OGBase~\cite{Li_2023_CVPR}             & \textcolor{blue}{2023}                & \multicolumn{4}{c|}{-}          & -          & 84.7          & 88.4          & 89.4          & -          & -           \\
                          & GaitBase~\cite{Fan_2023_CVPR}           & \textcolor{blue}{2023}                & 71.6          & 75.0          & 76.8          & 78.6          & 75.5          & 88.5          & 92.7          & 93.4         & 93.2          & 92.0           \\ \cmidrule{2-13}
& DeepGaitV2        & Ours                  & \textbf{78.6}          & \textbf{84.8}          & \textbf{80.7}          & \textbf{89.2}          & \textbf{83.3}          & \textbf{90.5} & \textbf{96.3} & \textbf{91.4}          & \textbf{96.7}          & \textbf{93.7}           \\ \midrule
\multirow{4}{*}{\begin{tabular}[c]{@{}c@{}}Skeleton-\\ based \end{tabular}} & GaitGraph2~\cite{teepe2021gaitgraph}        & \textcolor{blue}{2022}               & 5.0           & 5.3           & 5.8           & 6.2           & 5.1           & 5.0           & 5.7           & 7.3           & 8.8           & 6.7            \\
                          & Gait-TR~\cite{zhang2022spatial}           & \textcolor{blue}{2023}                  & 15.7          & 18.3          & 18.5          & 17.5          & 17.5          & 24.3          & 28.7          & 31.1          & 28.1          & 28.1           \\
                          & MSGG~\cite{peng2021learning}              & \textcolor{blue}{2023}                 & 29.0          & 34.5          & 37.1          & 33.3          & 33.5          & 43.1          & 52.9          & 57.4          & 49.9          & 50.8           \\ \cmidrule{2-13}
& SkeletonGait              & Ours                & \underline{40.4}     & \underline{48.5}   & \underline{53.0}     & \underline{61.7}     & \underline{50.9}      & \underline{52.4}    & \underline{65.4}     & \underline{72.8}   & \underline{80.9}  & \underline{67.9}  \\ \midrule
\multirow{2}{*}{\begin{tabular}[c]{@{}c@{}}Multi-\\ modal \end{tabular}} & BiFusion~\cite{peng2021learning}      & \textcolor{blue}{2023}         & 62.6     & 67.6   & 66.3     & 66.0     & 65.6      & 77.5    & 84.8     & 84.8   & 82.9  & 82.5  \\
& ParsingGait~\cite{zheng2023parsing}      & \textcolor{blue}{2023}     & 55.3    & 58.9     & 64.0   & 66.7  & 61.2  & 73.5     & 78.4   & 85.2     & 87.0     & 81.0     \\ \cmidrule{2-13}
& SkeletonGait++ & Ours                & \textbf{\underline{79.1}}          & \textbf{\underline{83.9}}          & \textbf{\underline{81.7}}          & \textbf{\underline{89.9}}          & \textbf{\underline{83.7}}          & \textbf{\underline{90.2}}          & \textbf{\underline{95.0}}          & \textbf{\underline{92.9}}          & \textbf{\underline{96.9}}          & \textbf{\underline{93.8}}  \\       
\bottomrule
\end{tabular}
\textit{
Note: this table presents the rank-1 accuracy.
}
\end{threeparttable}
\label{tab:results_on_CCPG}
\end{table*}

\begin{figure}[t]
\centering
\includegraphics[height=6.0cm]{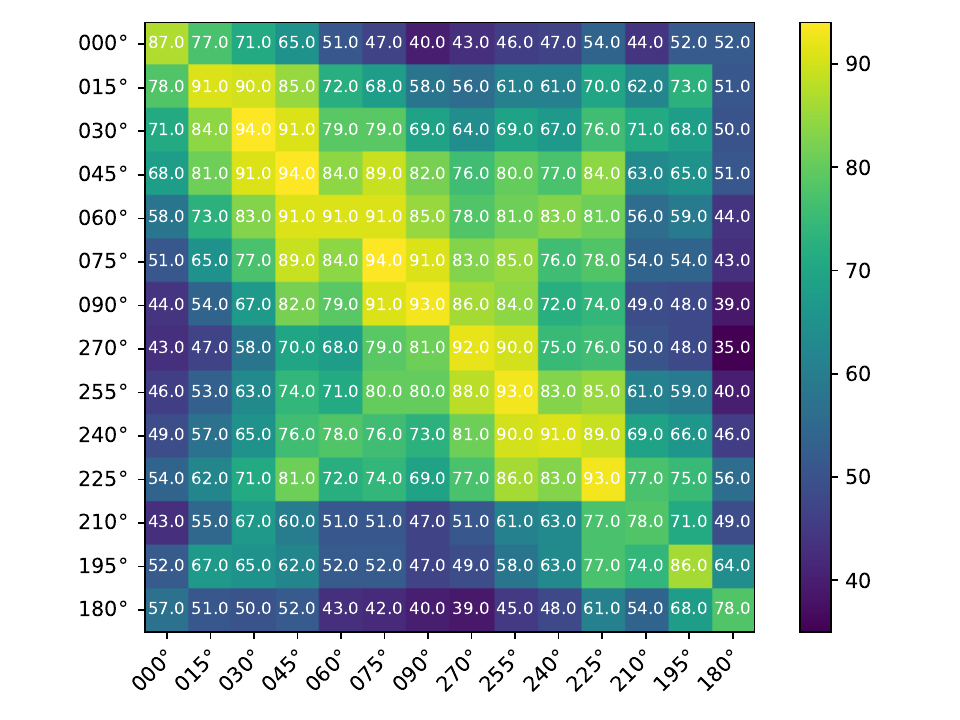}
\caption{
The accuracies of SkeletonGait over all probe-gallery view pairs on OU-MVLP. 
The horizontal and vertical coordinates respectively represent the filming viewpoints of the probe and gallery set. 
}
\label{fig:viewpoint}
\end{figure}

As discussed in the previous section, the skeleton map can be perceived as a silhouette devoid of body shape.
By comparing the rank-1 accuracy of DeepGaitV2 \textit{vs.} SkeletonGait in Table~\ref{tab:main_results}, \ref{tab:results_on_SUSTech1K}, and \ref{tab:results_on_CCPG}, \textit{i.e.}, 91.9\% \textit{vs.} 67.4\% on OU-MVLP~\cite{takemura2018multi}, 74.4\% \textit{vs.} 38.1\% on Gait3D~\cite{zheng2022gait3d}, 77.7\% \textit{vs.} 77.4\% on GREW~\cite{zhu2021gait}, 82.3\% \textit{vs.} 63.0\% SUSTech1K~\cite{Shen_2023_CVPR}, and 83.3\% \textit{vs.} 50.9\% on CCPG~\cite{Li_2023_CVPR}, 
we obtain the following thoughts:
\begin{itemize}
    \item \textbf{Importance of Structural Features}. 
    Structural features play an important role. 
    They contribute to over 50\% of the overall performance, as indicated by the rank-1 accuracy ratios between SkeletonGait and DeepGaitV2.
    \item \textbf{Superiority of Skeleton Data}. 
    In situations where silhouettes are noisy, such as the data at night in SUSTech1K, SkeletonGait exhibits a considerable performance advantage over DeepGaitV2 (50.0\% \textit{vs.} 28.8\%, Table~\ref{tab:results_on_SUSTech1K}). It shows the robustness of the skeleton data in such cases.
    \item \textbf{Cross-view Challenge}. The results on the diagonal and anti-diagonal of Fig.~\ref{fig:viewpoint} respectively are the rank-1 accuracies of SkeletonGait in identical- and nearly symmetric-view cases, where others reflect the cross-view situations on OU-MVLP~\cite{takemura2018multi}. As shown in the figure, the cross-view issue remains a major challenge for SkeletonGait. 
    \item \textbf{Concerns on GREW}. SkeletonGait cannot get comparable results as DeepGaitV2 on OU-MVLP and Gait3D. It just achieves 67.4\% (\textit{vs.} 91.9\% by DeepGaitV2) on OU-MVLP, and 38.1\% (\textit{vs.} 74.4\% by DeepGaitV2) on Gait3D according to Table~\ref{tab:main_results}. But surprisingly it can achieve a comparable accuracy of 77.4\% (\textit{vs.} 77.7\% by DeepGaitV2) on GREW.
    GREW has been widely acknowledged as the most challenging gait dataset due to its largest scale and real-world settings.
    Given that SkeletonGait performs well on the same view but poorly on cross views, as shown in the cross-view experiments on OU-MVLP (Fig.~\ref{fig:viewpoint}), one possibility is that GREW does not contain enough cross-view variations. However, further proofs are needed to rigorously confirm this point.
\end{itemize}

\subsection{Analysis by the Multi-modal SkeletonGait++}
As demonstrated in Table~\ref{tab:main_results}, \ref{tab:results_on_SUSTech1K}, and \ref{tab:results_on_CCPG}, SkeletonGait++ achieves a new SoTA with notable improvements compared to other single-modal and multi-modal methods.
Specifically, it gains $+1.4\%$, $+21.2\%$, $+9.5\%$, and $23.2\%$ rank-1 accuracy on the Gait3D~\cite{zheng2022gait3d}, GREW~\cite{zhu2021gait}, SUSTech1K~\cite{Shen_2023_CVPR}, and CCPG~\cite{Li_2023_CVPR} datasets, respectively. 
The result missing on OU-MVLP in Table~\ref{tab:main_results} is caused by the lack of frame-by-frame alignment between the skeleton and silhouette, which was not provided by the dataset owner and cannot be obtained by some methods. 

Compared to DeepGaitV2, the additional skeleton branch of SkeletonGait++ notably enhances the recognition accuracy, particularly when the body shape becomes less reliable. 
This augmentation is explicitly evident in challenging scenarios involving object carrying, occlusion, and poor illumination conditions, as observed on SUSTech1K (Table~\ref{tab:results_on_SUSTech1K}).

\begin{figure}[t]
\centering
\includegraphics[height=5.5cm]{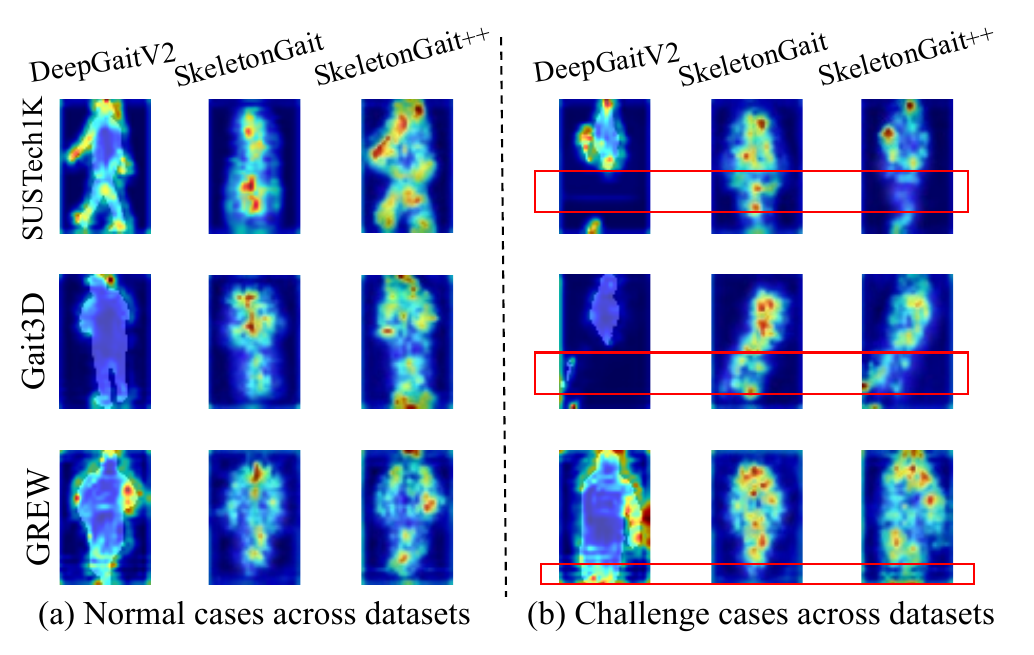}
\caption{
The activation visualization of DeepGaitV2 \textit{vs.} SkeletonGait and SkeletonGait++ (vertical axis) on some typical samples from SUSTech1K~\cite{Shen_2023_CVPR}, Gait3D~\cite{zheng2022gait3d}, and GREW~\cite{zhu2021gait} (horizontal axis).
}
\label{fig:visualization}
\end{figure}

To better understand the differences among DeepGaitV2, SkeletonGait, and SkeletonGait++, 
Fig.~\ref{fig:visualization} visualizes their activation maps through the CAM technique~\cite{zhou2016learning}. 
As a result, DeepGaitV2 pays more attention to regions that exhibit distinct and discriminative body shapes. 
On the other hand, SkeletonGait can only concentrate on \textit{clean} structural features over the body joints and limbs.
In comparison, SkeletonGait++ strikes a balance between these approaches, effectively capturing the \textit{comprehensive} gait patterns that are rich in both body shape and structural characteristics. 
Especially for the challenging cases shown in Fig.~\ref{fig:visualization} (b), SkeletonGait++ adaptively leverages the skeleton branch to support the robust gait representation learning. This is an urgent need for real applications. We also think it is the main reason for the improvement by SkeletonGait++ on Gait3D~\cite{zheng2022gait3d} and GREW~\cite{zhu2021gait}. 

\subsection{Ablation Study on Hyper-parameters}
The previous part of the section has thoroughly explored the relationship and distinction among DeepGaitV2, SkeletonGait, and SkeletonGait++, through comprehensive experiments across multiple popular gait datasets. 
The robustness of those methods to their hyper-parameters is analyzed below.

To determine a suitable network size for DeepGaitV2, the results by different depths (10, 14, 22, and 30 layers) are shown in Table~\ref{tab:overfitting}. 
We also changed the number of channels from 32 to 128, and the results are in Table~\ref{tab:ablation_study}~(a). Widen the network from 64 to 128 gains limited accuracy improvement but introduces significant training parameters and computation costs.
Therefore, we set the number of channels $C$ to 64 for a better accuracy-computation trade-off. 

To determine the empirical value of $\sigma$ in generating the skeleton map as described in Eq.~(\ref{equ:joints}) and (\ref{equ:limbs}), we tested it from 4 to 16, and the results in 
Table~\ref{tab:ablation_study}~(b) shows that: 
i) SkeletonGait is robust to the value of $\sigma$.
ii) $\sigma=8.0$ is an experimentally optimal choice.
These findings also underscore the robustness and discriminative capacity of the proposed skeleton map as a gait modality for gait recognition. 

To determine the fusion location and mode within SkeletonGait++ shown in Fig.~\ref{fig:skeletongait}, we tried all the combinations. The results in
Table~\ref{tab:ablation_study}~(c) reveals that: i) SkeletonGait++ is robust to both fusion locations and fusion modes. 
ii) The low-level attention fusion is an experimentally optimal choice. 
Moreover, these observations also highlight the promising potential of multi-modal gait recognition methods. 

\begin{table*}[htbp]
\caption{
Ablation Study on Gait3D~\cite{zheng2022gait3d}.
}
\begin{threeparttable}
    \begin{minipage}{0.30\linewidth}
        \centering
        \begin{tabular}{c|ccc}
        \multicolumn{4}{c}{(a) DeepGaitV2} \\
        \toprule
        Condition & \multirow{2}{*}{Rank-1} & \multirow{2}{*}{Param.} & \multirow{2}{*}{GFlops} \\
        $C$ in Table~\ref{tab:deepgaitv2} & & & \\ \hline \hline
        $C$=32        & 67.9    & 2.8M     & 0.7G     \\
        $C$=64        & 74.4    & 11.1M    & 2.9G     \\
        $C$=128       & \textbf{75.0}    & 44.4M    & 11.4G    \\
        \bottomrule
        \end{tabular}
    \end{minipage}%
    \quad \quad
    \begin{minipage}{0.15\linewidth}
        \centering
        \begin{tabular}{c|cc}
        \multicolumn{3}{c}{(b) SkeletonGait} \\
        \toprule
        \multicolumn{1}{c|}{Condition} & \multicolumn{2}{c}{Performance} \\ 
        $\sigma$ in Eq.~(\ref{equ:joints})                    & Rank-1  & mAP \\ \hline \hline
        $\sigma=4.0$                      & 37.5    & 28.5  \\
        $\sigma=8.0$                      & \textbf{38.1}    & \textbf{28.9} \\
        $\sigma=16.0$                     & 36.0    & 26.9 \\ \bottomrule 
        \end{tabular}
    \end{minipage}%
    \quad \quad
    \begin{minipage}{0.30\linewidth}
        \centering
        \begin{tabular}{c|cc|cc}
        \multicolumn{5}{c}{(c) SkeletonGait++} \\
        \toprule
        Condition & \multicolumn{2}{c|}{Low-Level} & \multicolumn{2}{c}{High-Level} \\
        Fusion in Fig.~\ref{fig:skeletongait}                      & Rank-1 & mAP  & Rank-1  & mAP  \\ \hline \hline
        Add                   & 76.5   & 69.6 & 76.2    & 69.5 \\
        Cat                   & 76.7   & 69.7 & 42.2    & 69.4 \\ 
        Attention             & 77.6   & \textbf{70.3}  & \textbf{78.2}  & 70.2 \\
        \bottomrule
        \end{tabular}
    \end{minipage}
\label{tab:ablation_study}
\vspace{0.1cm}
\textit{Note: the highest performance is in} \textbf{bold}. 
\end{threeparttable}
\end{table*}

\section{Trends and Challenges}
\label{sec:challenges}
\textcolor{blue}{
Despite significant efforts, we acknowledge that OpenGait has yet to address several crucial topics essential for advancing gait recognition toward broader real-world applicability.
To bridge this gap, this section offers key insights into the current state of gait recognition and outlines a clear roadmap for future advancements, highlighting emerging trends and opportunities.
Building on recent literature and feedback from the OpenGait community, we present the following discussion:
}

\noindent $\bullet$ 
\textcolor{blue}{
\textbf{Handling Noisy Inputs}. 
Most widely used gait input modalities, such as silhouettes and skeletons, are typically extracted from raw videos using third-party upstream algorithms, including pedestrian segmentation and pose estimation. 
However, accurately extracting these modalities from less-than-ideal environments pose their own set of challenges. 
Current approaches to mitigate this issue fall into three main directions:
(a) enhancing upstream models—though this falls outside the core scope of gait recognition research,
(b) exploring more robust modalities, such as point clouds~\cite{Shen_2023_CVPR},
(c) developing quality-robust gait models~\cite{hou2022gait, wang2024qagait}, 
(d) leveraging the complementarity of multi-modal inputs to enhance gait representation~\cite{xu2023occlussion}, and 
(e) facilitating end-to-end modeling~\cite{song2019gaitnet, li2020end, zhang2020learning, liang2022gaitedge, ye2024biggait} to eliminate the inaccuracy accumulation. 
However, a comprehensive evaluation comparing the effectiveness of these strategies remains lacking, leaving an open question in the field.
}

\noindent $\bullet$
\textcolor{blue}{
\textbf{Ideal Gait Input}. 
While RGB data can introduce distractions caused by variations in clothing, lighting, and backgrounds, they could also provide rich identity cues such as fine-grained body segmentation and 3D body priors.
To enhance gait representations, recent studies have explored:
(a) advanced sensors such as LiDAR for point cloud-based gait analysis~\cite{Shen_2023_CVPR, han2024gait}, 
(b) improved gait modalities, including human parsing~\cite{zheng2023parsing, zou2024cross} and SMPL models~\cite{li2020end, zheng2022gait3d}, 
(c) multi-modal frameworks integrating complementary gait representations~\cite{zheng2022gait3d, zou2024multi, fan2024skeletongait, jin2025exploring}, and
(d) end-to-end training strategies with inductive biases, such as appearance reconstruction~\cite{li2020end, zhang2020learning}, differentiable silhouette edge modeling~\cite{liang2022gaitedge}, and spatial feature smoothing~\cite{ye2024biggait}.
OpenGait recognizes that achieving an ideal gait input—one that is both rich in discriminative features and robust to irrelevant variations—remains an open challenge in the field.
}

\noindent $\bullet$ 
\textcolor{blue}{
\textbf{Real-world Challenges}. 
Despite significant data collection efforts, current gait recognition datasets still struggle to comprehensively capture real-world noise factors that influence walking patterns.
On one hand, constrained datasets~\cite{yu2006framework, takemura2018multi, Shen_2023_CVPR, Li_2023_CVPR, song2023casia, zou2024cross} allow for flexible annotation but is hard to replicate the complexities of real-world conditions, such as various occlusions, discontinuous videos, and illumination changes. 
On the other hand, in-the-wild datasets~\cite{zheng2022gait3d, zhu2021gait} better reflect practical challenges but face difficulties in accurately labeling variations within sequence pairs.
To mitigate these issues, recent studies have explored three key strategies:
(a) collecting datasets that focus on specific noise factors, such as different types of occlusions~\cite{huang2024occluded},
(b) leveraging pedestrian attribute classifiers to obtain automatic annotations for in-the-wild walking sequences~\cite{zhu2021gait}, and
(c) pretraining self-supervised gait models on large-scale unlabeled datasets to enhance robustness against noise factors~\cite{fan2022learning}.
OpenGait recognizes this as a critical direction in advancing gait recognition research toward better practicality.
}

\noindent $\bullet$ 
\textcolor{blue}{
\textbf{Gait Metric Learning}. 
Despite substantial accuracy gains across various gait benchmarks, the performance of existing models remains insufficient for real-world deployment.
Visual analyses of gait feature distributions~\cite{zheng2022gait, zheng2024takes} reinforce the observation that reducing intra-class variance remains a major challenge in gait recognition.
Beyond refining gait backbone architectures, recent research has explored two key strategies to address this issue:
(a) re-ranking techniques to improve retrieval accuracy by refining similarity relationships~\cite{wang2024free, habib2025cargait}, and
(b) enhanced loss functions tailored for optimizing gait metric learning~\cite{zhang2019learning}.
OpenGait recognizes gait metric learning as a crucial factor for improving real-world robustness, though it remains an area that has yet to receive sufficient research attention.
}

\noindent $\bullet$ 
\textcolor{blue}{
\textbf{Gait Temporal Modeling}. 
Previous literature widely recognizes two primary approaches to temporal modeling for gait description: set-based\cite{chao2019gaitset, hou2020gait, hou2021set} and sequence-based\cite{fan2020gaitpart, lin2021gait, huang2021context} methods.
In OpenGait, we also explore a set-based variant of DeepGaitV2, where the only modification is replacing ordered frames with shuffled frames during both training and testing.
This adjustment leads to a slight accuracy drop of no more than 5\% (70.2\% \textit{vs.} 74.4\%), aligning with similar findings in related studies~\cite{fan2020gaitpart, fan2023exploring}.
This suggests that holistic temporal dynamics across the entire frame set, rather than just local dependencies between neighboring frames, play a crucial role in existing methods.
Two possible explanations emerge:
(a) Holistic temporal dynamics may effectively capture gait details due to the periodic nature of gait cycles~\cite{chao2019gaitset}. 
(b) More advanced temporal models are still needed for a finer-grained description of gait motion.
OpenGait recognizes gait temporal modeling as an ongoing challenge in gait recognition research. 
}

\section{Conclusions}
With gait recognition becoming increasingly practicable, 
this work addresses the urgent need for developing a foundational gait codebase, revisiting previous state-of-the-art methods, and establishing new baselines. We provide a comprehensive benchmark study to encourage further advancements in the field. OpenGait offers an accessible platform, while DeepGaitV2, SkeletonGait, and SkeletonGait++ represent new baseline models for appearance-based, model-based, and multi-modal gait recognition, respectively.

Moreover, our work highlights three main challenges in gait recognition: dataset scale, model size, and modality design. 
Typically, indoor settings facilitate obtaining permission from volunteers for gait research but struggle to replicate real-world conditions. Conversely, outdoor gait datasets better capture real-world changes but present challenges in acquiring the walking crowd with consent. 
Therefore, the research community must reduce its heavy reliance on large-scale datasets. Emerging unsupervised and self-supervised methods offer promising solutions.
Building upon the dataset issues, our comprehensive experimental investigation addresses the problem of determining suitable capacity for deep gait models, exemplified by the DeepGaitV2 series.
Beyond only the appearance-based gait methodology, this paper further reveals the continued need for a discriminative yet clean gait modality by systematically comparing three proposed baseline models.
By highlighting these challenges and discussing emerging trends and opportunities, we hope to pave the way for future research aimed at improving real-world gait recognition applications.

\section{Acknowledgements}
We sincerely express our heartfelt gratitude to the OpenGait community for their invaluable support and collaboration. The insights provided by the Associate Editor and Reviewers of this paper have been instrumental in shaping Sec.~\ref{sec:challenges}, for which we are deeply thankful.
We extend our special thanks to Mr. Jilong Wang, Mr. Beibei Lin, Mrs. Yunjie Peng, Mr. Zirui Zhou, and Mr. Dingqiang Ye for their significant contributions of code to OpenGait.
We also deeply appreciate the support and extensions provided by the teams behind Gait3D~\cite{zheng2022gait3d}, GREW~\cite{zhu2021gait}, FastPoseGait~\cite{meng2023fastposegait}, CASIA-E~\cite{song2023casia}, CCPG~\cite{Li_2023_CVPR}, and CCGR~\cite{zou2024cross}, which have greatly enriched the field and our work.

This work is jointly supported by the National Natural Science Foundation of China (62476120, 62206022, 62276025, 62476027),  Shenzhen International Research Cooperation Project (GJHZ20220913142611021), Beijing Municipal Science \& Technology Commission (Z231100007423015), and Scientific Foundation for Youth Scholars of Shenzhen University (868-000001033383).

We would like to thank Mr. Chunhua Wang for his professional technical support to our work.

\bibliographystyle{IEEEtran}
\bibliography{egbib}

\newpage
\begin{IEEEbiography}[{\includegraphics[width=1in,height=1.25in,clip,keepaspectratio]{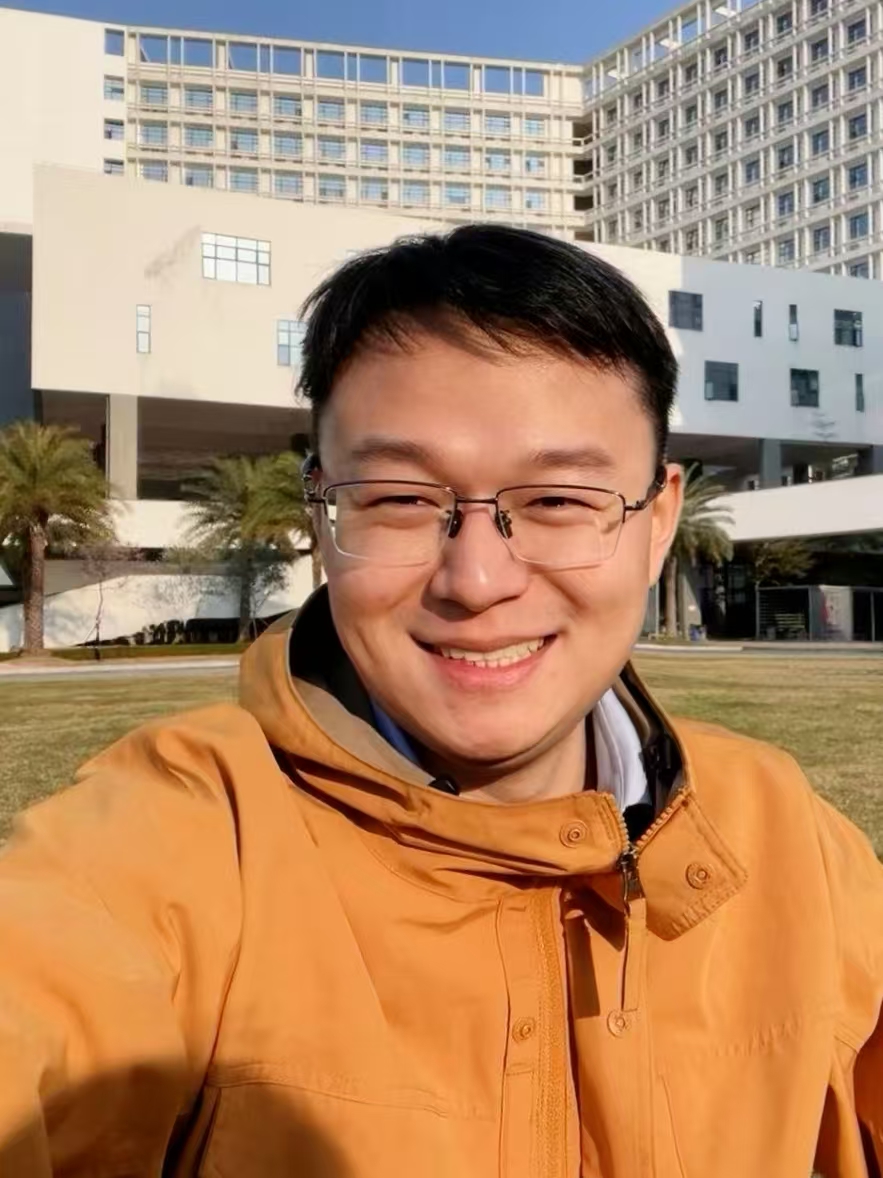}}]{Chao Fan} received his B.E. degree from Xi’an University of Technology in 2018, his M.S. degree from the University of Science and Technology Beijing in 2021, and his Ph.D. from the Southern University of Science and Technology in 2024. He is currently an Assistant Professor at the School of Artificial Intelligence, Shenzhen University (SZU), and also at the National Engineering Laboratory for Big Data System Computing Technology, SZU. 
His research interests include computer vision and intelligent perception.
\end{IEEEbiography}

\begin{IEEEbiography}[{\includegraphics[width=1in,height=1.25in,clip,keepaspectratio]{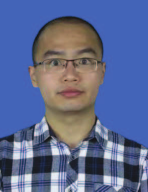}}]{Saihui Hou}
received the B.E. and Ph.D. degrees from the University of Science and Technology of China in 2014 and 2019, respectively.
He is currently an Associate Professor at the School of Artificial Intelligence, Beijing Normal University.
His research interests include computer vision and machine learning, with a focus on gait recognition, which aims to identify individuals based on their walking patterns.
\end{IEEEbiography}

\begin{IEEEbiography}[{\includegraphics[width=1in,height=1.25in,clip,keepaspectratio]{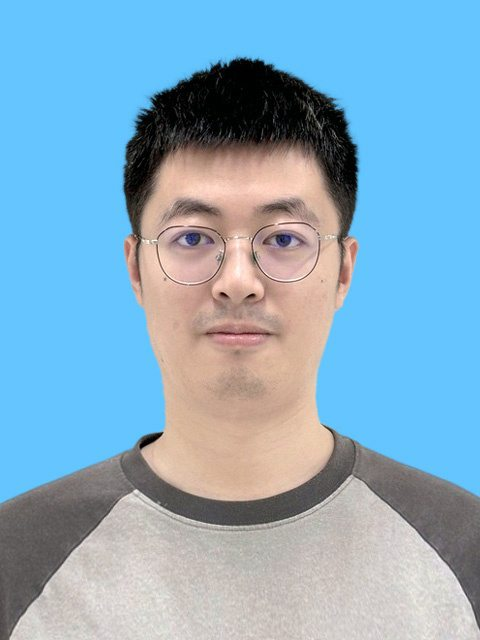}}]{Junhao Liang} graduated from Southern University of Science and Technology, obtaining his B.E. in 2021 and M.E. in 2024, both from the Department of Computer Science and Engineering. Currently, he is working at the Department of Computer Vision Technology (VIS), Baidu Inc., focusing on AIGC related to digital humans. 
\end{IEEEbiography}

\begin{IEEEbiography}[{\includegraphics[width=1in,height=1.25in,clip,keepaspectratio]{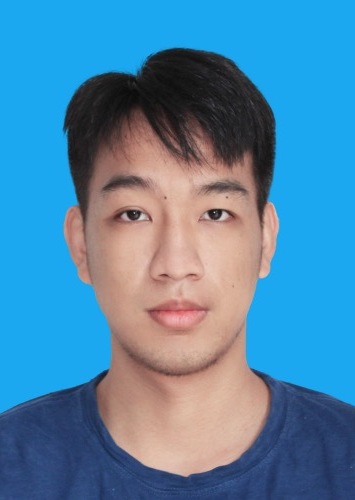}}]{Chuanfu Shen}
is a postdoctoral researcher at the Shenzhen Institute of Advanced Study, University of Electronic Science and Technology of China (UESTC). He received his B.E. degree from the Southern University of Science and Technology in Shenzhen, China, and obtained his Ph.D. jointly enrolled in Computer Science and Engineering at the Southern University of Science and Technology, and Data and Systems Engineering at The University of Hong Kong. His research interests include human retrieval, gait recognition, and 3D perception.
\end{IEEEbiography}

\begin{IEEEbiography}[{\includegraphics[width=1in,height=1.25in,clip,keepaspectratio]{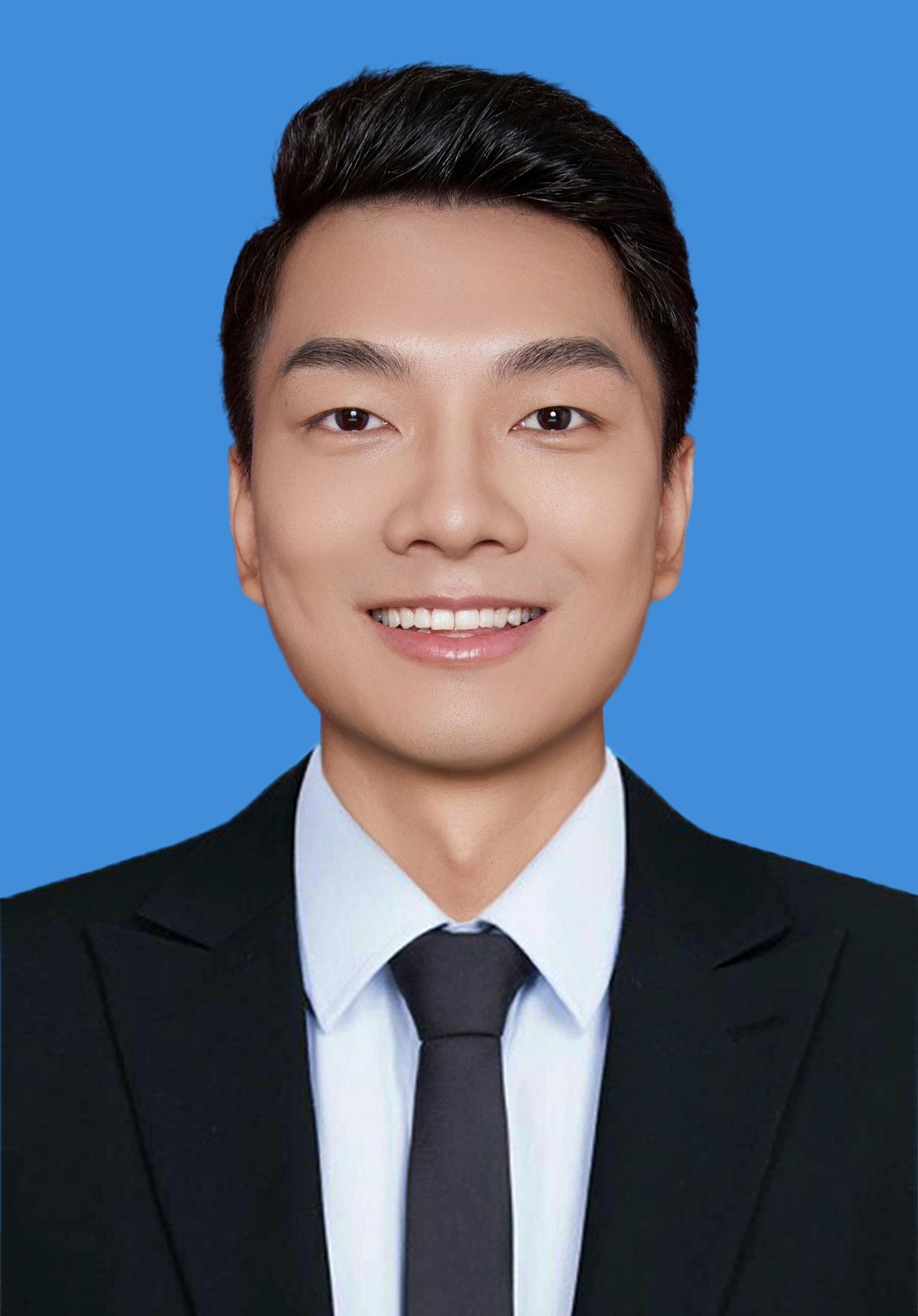}}]{Jingzhe Ma}
received the B.E. and M.S. degrees in Computer Science and Technology from Zhengzhou University in 2017 and 2020, respectively, and the Ph.D. degree from the Department of Computer Science and Engineering at Southern University of Science and Technology in 2024. He is currently a Research Assistant Professor at Institute of Applied Artificial Intelligence of the Guangdong-Hong Kong-Macao Greater Bay Area, Shenzhen Polytechnic University. His research interests include human video synthesis, gait recognition, and gait anonymization.
\end{IEEEbiography}

\begin{IEEEbiography}[{\includegraphics[width=1in,height=1.25in,clip,keepaspectratio]{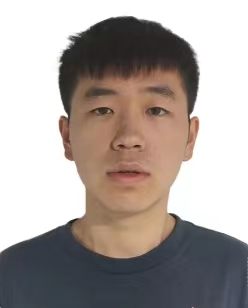}}]{Dongyang Jin}
received the B.E. degree from Southern University of Science and Technology in 2023. He is currently a Master student with Department of Computer Science and Engineering, Southern University of Science and Technology. His research interests include gait recognition and AIGC.
\end{IEEEbiography}
\vspace{-9cm}

\begin{IEEEbiography}[{\includegraphics[width=1in,height=1.25in,clip,keepaspectratio]{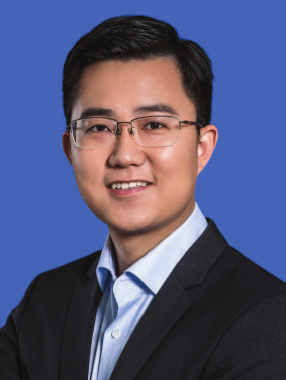}}]{Yongzhen Huang}
 received the B.E. degree from Huazhong University of Science and Technology in 2006, and the Ph.D. degree from Institute of Automation, Chinese Academy of Sciences in 2011.
 He is currently a Professor at School of Artificial Intelligence, Beijing Normal University.
 He has published one book and more than 120 papers at international journals and conferences such as TPAMI, IJCV, TIP, TSMCB, TMM, TCSVT, CVPR, ICCV, ECCV, NIPS, AAAI.
 His research interests include pattern recognition, computer vision and machine learning.
\end{IEEEbiography}
\vspace{-9cm}

\begin{IEEEbiography}[{\includegraphics[width=1in,height=1.25in,clip,keepaspectratio]{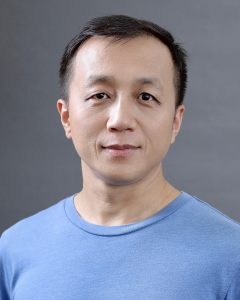}}]{Shiqi Yu} is currently an Associate Professor in the Department of Computer Science and Engineering, Southern University of Science and Technology, China. He received his B.E. degree in computer science and engineering from the Chu Kochen Honors College, Zhejiang University in 2002, and Ph.D. degree in pattern recognition and intelligent systems from the Institute of Automation, Chinese Academy of Sciences in 2007. He worked as an Assistant Professor and an Associate Professor in Shenzhen Institutes of Advanced Technology, Chinese Academy of Sciences from 2007 to 2010, and as an Associate Professor in Shenzhen University from 2010 to 2019. His research interests include gait recognition, face detection and computer vision.
\end{IEEEbiography}

\end{document}